\documentclass[preprint,12pt]{elsarticle} 

\usepackage{amssymb}
\usepackage{amsmath}

\usepackage{url}
\usepackage{multirow} 
\usepackage[shortlabels]{enumitem}
\usepackage{setspace}

\journal{Sustainable Energy, Grids and Networks}

\begin{document}

\begin{frontmatter}
\title{Extending Load Forecasting from Zonal Aggregates to Individual Nodes for Transmission System Operators}

\author[SU]{Oskar Triebe\corref{corr}}
\cortext[corr]{triebe@stanford.edu}
\author[SU]{Fletcher Passow}
\author[TUM]{Simon Wittner}
\author[TUM]{Leonie Wagner}
\author[TUM]{Julio Arend}
\author[SU]{Tao Sun}
\author[SU]{Chad Zanocco}
\author[CTU]{Marek Miltner}
\author[MISO]{Arezou Ghesmati}
\author[MISO]{Chen-Hao Tsai}
\author[Granada,Monash]{Christoph Bergmeir}
\author[SU]{Ram Rajagopal}

\affiliation[SU]{
organization={School of Engineering, Stanford University}, 
state={California},
country={United States}
}
\affiliation[MISO]{
organization={Midcontinent Independent System Operator}, 
country={United States}
}
\affiliation[TUM]{
organization={Technical University Munich}, 
country={Germany}
}
\affiliation[CTU]{
organization={Czech Technical University in Prague}, 
country={Czech Republic}
}
\affiliation[Granada]{
organization={University of Granada}, 
country={Spain}
}
\affiliation[Monash]{
organization={Monash University}, 
country={Australia}
}

\begin{abstract}
The reliability of local power grid infrastructure is challenged by sustainable energy developments increasing electric load uncertainty. Transmission System Operators (TSOs) need load forecasts of higher spatial resolution, extending current forecasting operations from zonal aggregates to individual nodes. However, nodal loads are less accurate to forecast and require a large number of individual forecasts, which are hard to manage for the human experts assessing risks in the control room's daily operations (operator). In collaboration with a TSO, we design a multi-level system that meets the needs of operators for hourly day-ahead load forecasting. Utilizing a uniquely extensive dataset of zonal and nodal net loads, we experimentally evaluate our system components. First, we develop an interpretable and scalable forecasting model that allows for TSOs to gradually extend zonal operations to include nodal forecasts. Second, we evaluate solutions to address the heterogeneity and volatility of nodal load, subject to a trade-off. Third, our system is manageable with a fully parallelized single-model forecasting workflow. Our results show accuracy and interpretability improvements for zonal forecasts, and substantial improvements for nodal forecasts. In practice, our multi-level forecasting system allows operators to adjust forecasts with unprecedented confidence and accuracy, and to diagnose otherwise opaque errors precisely. 
\end{abstract}




\end{frontmatter}



\section{Introduction}
Electric transmission system operators (TSOs) face increasing volatility in electric load due to distributed and renewable energy generation, climate events, and electrification \cite{gielen_role_2019}.
This volatility complicates load forecasting, which is essential to TSO operations. TSOs must ensure that electricity generation matches load at all times, and the distribution of power across their territory does not overwhelm any infrastructure component. To accomplish this, they use day-ahead load forecasts to inform where to dispatch generators each hour of the coming day. Growing electrification and distributed generation increase volatility of `net load' -- local consumption minus generation -- in some places and not others, as adoption of these technologies proceeds unevenly. This could put a TSO's medium-voltage grid components, for example sub-transmission lines and primary distribution substations, at risk of damage if load forecasts miss unexpected local changes \cite{lv-review, wang_geospatial_2023,substation-storage-5-outof-56}.
To manage local volatility, TSOs need day-ahead net load forecasts with high spatial resolution which are also manageable and interpretable by human experts. 
However, existing solutions in academia and practice do not meet one or more of these criteria. 
Taking a TSO's perspective, we experimentally uncover the challenges and benefits of integrating high-resolution forecasts.
We contribute an interpretable model and a multi-level forecasting approach which we validate on an extensive TSO dataset.

\section{Problem Definition} 
Many TSOs have started collecting nodal load data in the last two decades, providing the training data needed to increase spatial resolution. 
However, high-spatial-resolution load time series are inherently challenging to forecast due to their volatility and heterogeneity.
We categorize aggregation levels of spatial resolution following ANSI C84.1-2020:
\textit{`system level'} aggregate power demand across a TSO's entire territory within their balancing mandate, \textit{`utility level'} aggregate regional power demand of individual utility territories or climate zones, related to high-voltage transmission systems  (115 kV and up, GW power range), \textit{`bus level'} medium-voltage load at subtransmission system nodes and primary substations (4 kV to 69 kV, MW power range), and \textit{`distribution level'} low-voltage load at secondary distribution systems (2.4 kV and below, kW power range). 

\subsection{Short-Term Bus Load Forecasting}

As spatial resolution increases, load forecasts become less precise and may require specialized models due to increased volatility and heterogeneity \cite{sun2013efficient, tao-hong-smart-meter-agg}. This deterioration accelerates when increasing spatial resolution beyond the bus level. Increasing resolution to loads below 1 MW leads to a steep decline in accuracy, whereas reducing resolution to larger aggregates of load increases accuracy at a slower rate \cite{ram-scaling-law}.  
This positions the bus level as an important forecasting level not only from a power systems perspective, but also regarding the accuracy-resolution trade-off. 
While bus loads constitute a an ideal forecasting level, their resolution is already high enough to be subject to vastly increased volatility and heterogeneity compared to the utility level, posing challenges to model selection and training.

To date, the majority of research work focuses on either low-resolution utility-level forecasts or extremely-high-resolution distribution-level forecasts \cite{lv-review}. 
Work on bus-level forecasting is very limited \cite{lv-review} and public bus-level datasets are lacking \cite{abdolrezaei2022substation}. Existing work shows promising performance \cite{sun2013efficient, CHEN2025epsr}, but is limited in applicability due to lack of interpretability and evaluation on few buses, with typical evaluations covering 1 to 3 areas or substations \cite{tampa-1-campus, map-reduce-2-series, gru-stlf-2-areas, 1996-ann-3-sub}, 9 substations \cite{nine-substations, transformer-nz}, 16 buses \cite{CHEN2025epsr}, 23 substations \cite{sun2013efficient}.
Other related work models the impact of home energy storage on the forecast error at five primary substations \cite{substation-storage-5-outof-56}, or proposes solutions for long-term forecasting of medium-voltage and high-voltage power demand ~\cite{ltlf-spatial, mtlf-feeder}.

Overall, bus-level forecasting work is lacking TSO relevance due to lack of operational integration, with evaluation scopes falling short of covering the full set of buses of an entire utility's territory. Additionally, TSOs require interpretability and manageability of their forecasting systems, discussed next.

\subsection{Forecast Interpretability} \label{interpretability-requirements}
In the TSO control room's daily operations, a human expert (`operator') assesses situational risks and adjusts forecasts to correct for internal biases of a model and external influences which a model may not take into account.
Sometimes, operators must do this in high-stress scenarios with limited time and high stakes, requiring a high level of skill and intuition.
To adjust the forecast confidently -- or decide not to adjust -- operators require forecast interpretability composed of: 
\begin{enumerate}[a),topsep=4pt,parsep=0pt,leftmargin=24pt]
\item A simple, intuitively understandable model \cite{chen_interpretable_2023}. 
\item Contextual reasoning about a specific situation -- a clear input-output mapping -- quantifying the effect of input features on forecasts \cite{chen_interpretable_2023}. 
\item Probabilistic forecasts quantifying the situation-specific uncertainty \cite{pinheiro2023short}.
\end{enumerate}

The use of a conditional uncertainty estimation method is necessary due to load uncertainty varying based on factors such as the time of year, weather conditions, and recent demand patterns.
A frequently used interpretable model type is a general additive model (GAM), composed of multiple individually interpretable components \cite{Hastie2017GeneralizedModels}. 
Deep learning methods -- while increasingly popular among researchers -- offer limited interpretability \cite{chen_interpretable_2023}, disqualifying many state-of-the-art models. 

\subsection{System Manageability}
High-resolution forecasting can become too complex for human operators to manage and interpret due to the number of forecasts, models and levels.
The effective complexity can be limited by choosing to monitor utility-level forecasts only by default.
This requires coherence across levels and a hierarchical reconciliation method, such as bottom-up aggregation \cite{hyndman2018}.

Another challenge to system manageability is the number of models in deployment. Currently, many TSOs fit one model per each utility load series. Applying this methodology at the bus level would increase the number of models by more than hundredfold, posing significant computational and model interpretation challenges.
However, recent advancements in machine learning techniques and global forecasting models (GFM), make it possible to fit a single model to many series producing individual and independent forecasts. GFMs are less susceptible to data volatility, but struggle with data heterogeneity \cite{hemawalage-gfm}.

The vast majority of existing short-term load forecasting research work prioritizes method novelty and accuracy on limited datasets over the outlined TSO requirements. 
More work is needed -- taking a TSO-centric approach -- towards bridging the gap between academic state-of-the-art high-resolution forecasting methods and actual TSO practices.

\subsection{Our Contribution}
This study represents a step towards the integration of bus load forecasts with existing TSO operations, not covered by previous work. 
The geographic scope of our study is uniquely extensive, covering an entire TSO system with individual load forecasts for all its constituent utility territories.
We evaluate the benefits of individually forecasting each of the constituent buses of an entire utility territory, with a coherent utility-level integration. Our contributions include:

\begin{enumerate}[1.,topsep=4pt,parsep=0pt,leftmargin=24pt]

    \item We develop an interpretable model that effectively scales to large numbers of forecasted time series. Our approach is suitable for both utility and bus level forecasting, which would allow a TSO to first transition their existing operations to our model with minimal workflow changes, and then gradually expand to bus-level forecasts with the same model.  
    
    \item We evaluate different hierarchical reconciliation and model training paradigms for their effectiveness to address heterogeneity and volatility challenges posed by bus load.
    
    \item We demonstrate how our model and multi-level approach equip TSO operators with powerful new tools to enable precise forecast adjustments and error source diagnostics.
    
\end{enumerate}

\section{Methodology} \label{sec:methodology}
Herein we describe the dataset, forecasting system, and evaluation setup.

\subsection{Dataset} \label{sec:dataset}
\begin{figure*}[htbp]
\centering
\includegraphics[width=\textwidth]{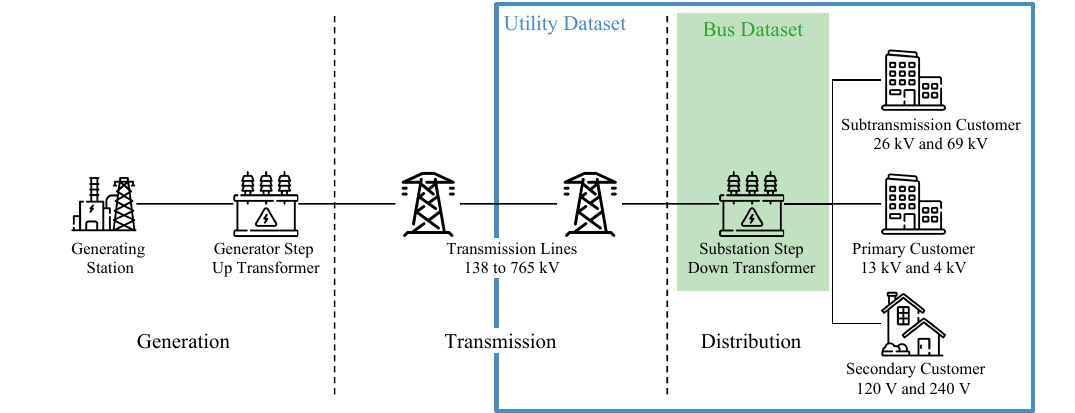}
\caption{Scope of the datasets visualized with a traditional power grid structure.}
\label{fig:data_description}
\end{figure*}

This study employs an extensive two-level TSO dataset 
from Midcontinent Independent System Operator (MISO)\footnotemark{}, measuring hourly net load demand from January 2018 to September 2021, illustrated in Figure~\ref{fig:data_description}.  
\footnotetext{
MISO is a TSO overseeing one of North America's most extensive wholesale electricity markets serving an estimated 42 million residents across 15 US states. The serviced territory includes diverse climate zones from the Gulf of Mexico to continental Canada.} 
The utility level dataset features the zonal power demand of each of the 37 MISO utilities. This power demand is calculated by subtracting the power exported from the power generated within that zone. For one anonymous utility within the utility dataset, the bus level dataset records the net load of each bus. Every bus in the utility is included, encompassing an entire aggregation zone with more than 100 buses in total. Average hourly bus loads range from 0.16~MWh to 37~MWh. To preserve the privacy of the loads and generators at these buses and the name of the utility, the exact number of buses within that utility is kept private. Hereafter, we refer to the bus dataset's utility as ``Utility A".
Supplementary Material \ref{appendix-methods} offers a more details on the datasets.

\subsection{Forecasting System}
We evaluate the essential building blocks of a multi-level forecasting system and their composition against the goals of a TSO operator. These blocks include a hierarchical reconciliation method, a data pooling approach for model training, and an underlying forecasting model type.

\subsubsection{Hierarchical Reconciliation Methods}
Today, TSO operators use utility level forecasts. Any proposed bus level forecasts thus need to be coherently integrated with the utility level. 
We evaluate two hierarchical reconciliation strategies between utility and bus level forecasts in Section \ref{sec:case_study}. Following a ``top-down" approach, utility level loads are forecasted and then split into bus level forecasts using historical proportions. Following a ``bottom-up" approach, bus level loads are directly forecasted, and then summed into utility level forecasts \cite{hyndman2018}.
All evaluated approaches directly forecast only one level, avoiding incoherent predictions that would require advanced hierarchical reconciliation \cite{hyndman2018}.

\subsubsection{Data Pooling Approaches for Model Training}\label{sec:global-local}
At the bus level, the number of series to forecast can become very numerous. In our study, we have more than 100. A traditional, ``local", univariate time series forecasting approach is to fit one model per each time series. For a large grid this could become overwhelming to TSO operators. In contrast, a ``global" forecasting model (GFM) follows the approach of using one model to forecast multiple univariate time series. A GFM is the time-series equivalent of a pooled regression model. 
At inference time, a GFM is indistinguishable from a local model. GFMs are univariate models, not to be confused with multivariate models, hierarchical methods, or time series aggregation.\footnote{
Global forecasting models are a special form of univariate models, different from multivariate models that model interaction effects, also not to be confused with the recently popularized foundation models. Foundation models are fitted on massive datasets of unrelated time series -- not reliably modeling the actual data at hand, and facing challenges in evaluation due to data leakage from pre-training. Foundation models furthermore rely on very large deep learning models that are difficult to interpret and adjust.}  
The only difference is in the training paradigm. Instead of fitting one model for one series, one model is fitted across the pool of all series to fit generally to all series. For a simple linear model, the resulting GFM is comparable to averaging the parameters of a set of local models. However, as our model has multiple jointly fitted non-linear components and is fitted via mini-batch stochastic gradient descent (SGD), the GFM training paradigm is not only more computationally efficient but also allows to fit the model components to a better global optimum with respect to its parameters.
At inference time, the GFM is applied to each series individually, producing independent forecasts for each series without interactions, akin to a local univariate model. In practice, we use parallelization to forecast all series at once.

In our study, we only pool time series \textit{within} a grid level. For example, we may pool bus load series together or utility load series together, but we never pool a mix of bus series and utility series together.   From a system manageability perspective, global methods have an advantage over local methods because they only require a single set of fitted parameters instead of one set of fitted parameters per series. This pooling further helps address volatility as it has a regularizing effect, cross-learning general patterns across all time series \cite{hemawalage-gfm}. 

\paragraph{Grouping Buses} \label{sec:grouping-buses}
Global modeling assumes homogeneity across time series. Thus, limiting a system to a single global model may be slightly too restrictive, since there can be substantial heterogeneity between buses. To allow for a compromise between the one-model-for-all-series of global methods and the one-model-per-series of local methods, we apply one global model to each \textit{group} of series in Section \ref{sec:case_study}.
We employ a grouping approach for the `Grouped Global-Bus' experiment in Section \ref{sec:case_study} .
Hereby, the objective is to mitigate the heterogeneity inherent to the bus dataset by grouping without trading off the volatility mitigation benefits of global models. 
The bus load series are clustered using a k-means algorithm into three groups.
The clustering features are composed of common time series features, i.e. trend, spike, linearity, curvature, stability, lumpiness, seasonal strength, trough, entropy, ACF1, and ACF10 \cite{Hyndman2020Time1.0.2}, described in Supplementary Material  \ref{appendix-methods}. 
As overfitting due to bus load volatility can become an issue if group sizes become small, we limit the number of groups to three. This allows for sufficiently large group sizes, where each model is expected to generalize well due to the regularizing effect of global modeling. 

\subsubsection{Forecasting Models}\label{sec:proposed_model}
First, we explain our proposed model, which is purpose-built to meet TSO operators' needs for interpretability and to be compatible with multiple data pooling approaches. 
Then, we describe our benchmark models we use in our utility-level investigation (see Section \ref{sec:results-utility}).

\begin{figure*}[!htb]
\centering
\includegraphics[width=\textwidth]{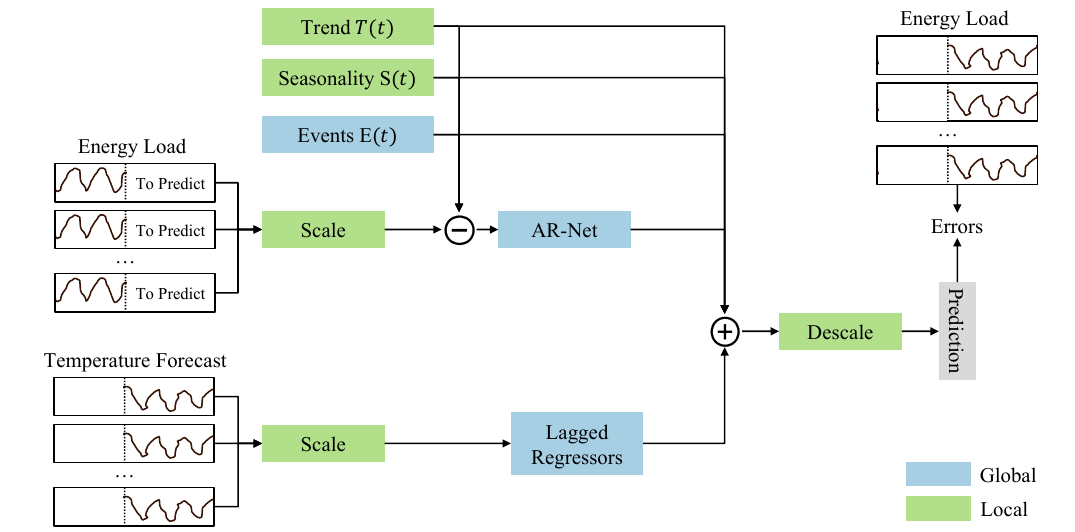}
\caption{
Architecture of hits-GAM: Trend, Seasonality and Event effects are computed for a given time and subtracted from the scaled load to yield a semi-stationary series, which is subsequently modeled via an autoregressive neural network. Concurrently, temperature forecasts are scaled and modeled with an independent neural network.
Finally, all components are summed up and re-scaled to produce the forecast. 
}
\label{fig:hitsGAM}
\end{figure*}

\paragraph{Proposed model} For the trained model at the foundation of our forecasting system, we propose the hybrid interpretable time series General Additive Model (hits-GAM), illustrated in Figure~\ref{fig:hitsGAM}. The model is hybrid in two ways: It fuses traditional and deep learning methods, as well as local and global modeling. 
The resulting model is highly interpretable, scales to forecast many series at once, and, as we shall demonstrate in later sections, produces accurate forecasts on both the utility and bus levels. 
The model is composed of additive components, each designed to capture different load dynamics:

\begin{itemize}[-,topsep=0pt,parsep=0pt,leftmargin=18pt]
\item \textit{Trend:} Captures slow changes and the long-term trend of energy load magnitude, using piece-wise linear sections with changepoints. 
\item \textit{Seasonality:} Recurring smooth seasonal patterns over typical time periods (daily, weekly or yearly), using a combination of Fourier terms.
\item \textit{Autoregression:} The observed load patterns of the last 15 days are regressed on the day ahead, using an autoregressive neural network \cite{Triebe2019AR-Net:Time-series}. 
\item \textit{Lagged Regressors:} Estimates the impact of forecasted temperatures, using an independent neural network. Note that the inputs are shifted such that this component regresses on the \textit{future} temperature forecast.
\item \textit{Events:} Estimates the offset effects of U.S. holidays. 
\end{itemize}

This modular approach retains the intuitive interpretability of individual components, despite jointly fitting all components using mini-batch Gradient Descent.  
For each component a suitable accuracy-interpretability trade-off \cite{chen_interpretable_2023} can be chosen (e.g. autoregressive neural networks for temperature, Fourier terms for seasonality). 
Furthermore, each component can be fitted locally or globally.
To reduce the risk of overfitting, we chose to fit the more expressive model components (autoregression and lagged regressors) in a global fashion, while the simpler model components (trend and seasonality) are fitted in a local fashion, as highlighted in Figure \ref{fig:hitsGAM}.
Additionally, a conditional forecast uncertainty estimate is provided using Quantile Regression \cite{Koenker1978RegressionQuantiles} with a Pinball Loss Function \cite{Steinwart2011EstimatingLoss}. 
We compose and manage the model using the open-source forecasting framework NeuralProphet\footnote{
The source code is available at the GitHub repository: \\\url{https://github.com/ourownstory/neural_prophet}} \cite{Triebe2021NeuralProphet:Scale}, shown to be suitable for interpretable forecasting \cite{Stromer2023DesigningForecasts}.

\paragraph{Comparison Models}\label{sec:baseline_models}
In the Utility-Level Forecasting Results section (Section \ref{sec:results-utility}), we assess the efficacy of the proposed hits-GAM by benchmarking with four popular forecasting models: Seasonal Autoregressive Integrated Moving Average (ARIMA) \cite{hyndman2018}, k-Nearest Neighbors (kNN) \cite{knn-load}, seasonal Na\"ive (sNa\"ive) \cite{hyndman2018}, and Extreme Gradient Boosting (XGBoost) \cite{Chen2016XGBoost:System}.
The traditional ARIMA, kNN, and sNa\"ive models are fitted locally, whereas XGBoost and hits-GAM are fitted as global models.
These comparison models were trained on the utility dataset using the same input data and equivalent hyperparameters as the hits-GAM model. 
The models, hyperparameters, and computational infrastructure are detailed in Supplementary Material \ref{appendix-methods}.

\begin{figure*}[htbp]
\centering
\includegraphics[width=\textwidth]{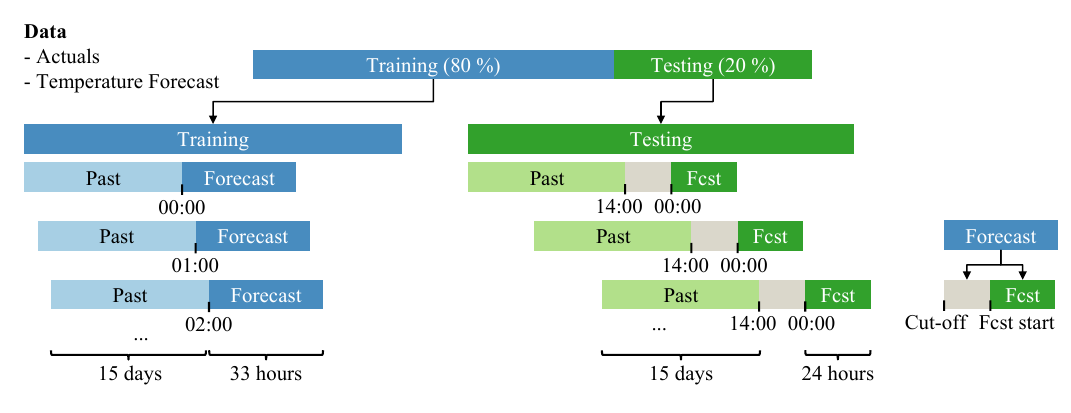}
\caption{Forecast origin for training and testing using a rolling window approach.}
\label{fig:flowchart}
\end{figure*}

\subsection{Experimental Evaluation and Metrics}\label{sec:experiments}
TSOs' day-ahead forecasts are computed on the day before the day being forecasted. We call these the `compute day' and the `target day', respectively. 
We follow MISO's practices for tracking day-ahead forecast performance, selecting the forecast produced at 2 PM on the compute day to evaluate the forecast accuracy on the target day.
The model is configured to forecast the next 33 hours, visualized in Figure \ref{fig:flowchart}. 
At 2pm of the compute day, the forecast covers the remaining 9 hours of the compute day and the 24 hours of the target day. We evaluate the model's performance only on the target day hours. 
In contrast, during training, a rolling window approach is employed with a one-hour step size, producing forecasts at each hour of the day. Hereby, all 33 hours of the forecasts are used to calculate errors and fit the model. This approach regularizes the model fit and uses all of the training data effectively.

The datasets of both levels are split to utilize 80\% for training (2018 through 2020), and 20\% for evaluation (first nine months of 2021).
Forecast accuracy is measured using error metrics popularized by the well recognized M competitions \cite{Makridakis2020}, i.e. Mean Absolute Error (MAE), Root Mean Square Error (RMSE), Mean Absolute Percentage Error (MAPE), Mean Absolute Scaled Error (MASE), and Mean Squared Scaled Error (MSSE). 
The scale-independent metrics MASE and MSSE enable a comparison across time series of different magnitudes \cite{Rostami-TabaraHierarchicalServices}. Hereby, the forecast error is scaled by the forecast error of a baseline method, i.e. 48-hour sNa\"ive in our study \cite{hyndman2018}. 
Due to many bus load observations close to zero, MAPE is not used in the bus level extension.
The error metric definitions are provided in the Supplementary Material  \ref{appendix-methods}.


\section{Utility-Level Forecasting Results} \label{sec:results-utility}
In order to ensure compatibility with current zonal aggregate forecasting operations, we first evaluate the accuracy and interpretability of our forecasting model, hits-GAM, at the utility level. We present an extension to the bus level in Section~\ref{sec:case_study}.

\begin{table}[htbp]\textbf{}
\centering
\caption{Average Utility-Level Forecast Accuracy}
\label{tab:weighted_avg}
\small
\begin{tabular}{lrrrrr}
\hline
Model & RMSE & MAE & MAPE & MASE & MSSE \\
\hline
ARIMA & 205.90 & 155.53 & 7.35 & 0.77 & 0.59 \\
kNN & 156.72 & 114.30 & 5.76 & 0.61 & 0.40 \\
sNa\"ive & 251.09	& 189.84 & 9.46 & 1.00 & 1.00 \\
XGBoost & 144.92 & 108.61 & 5.53 & 0.59 & 0.37 \\
hits-GAM & \textbf{105.41} & \textbf{78.22} & \textbf{4.21} & \textbf{0.58} & \textbf{0.36} \\ 
\hline
\end{tabular}
\end{table}

Overall, hits-GAM exhibits the highest predictive accuracy while also offering strong interpretability.
Our model shows large improvements in absolute error metrics (RMSE and MAE), with marginal improvements for scaled metrics (MASE and MSSE), presented in Table \ref{tab:weighted_avg}.
A more granular breakdown of the metrics is available in Supplementary Material \ref{appendix-results}.

\begin{figure}[!htbp]
\centering
\includegraphics[width=0.75\textwidth]{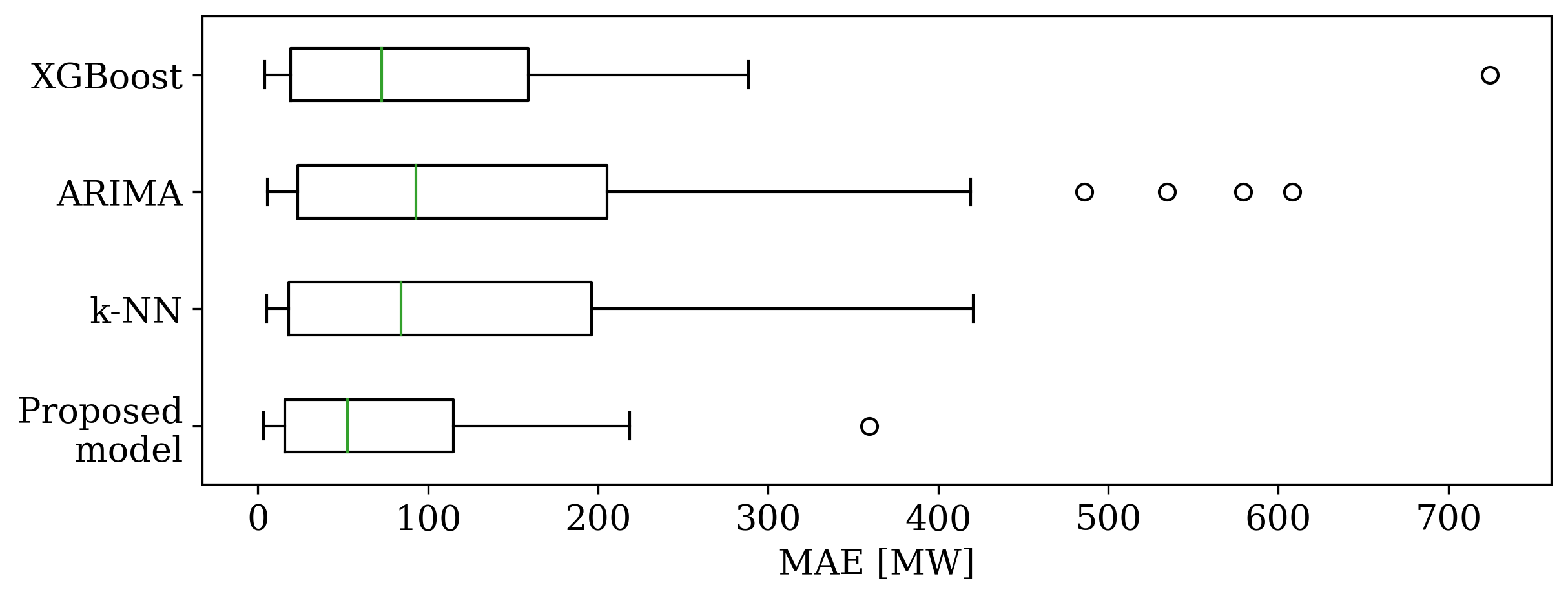}
\caption{Error distribution across utilities for utility-level MAE.}
\label{fig:error_distribution}
\end{figure}

\begin{figure}[!htb]
\centering
\includegraphics[width=0.75\textwidth]{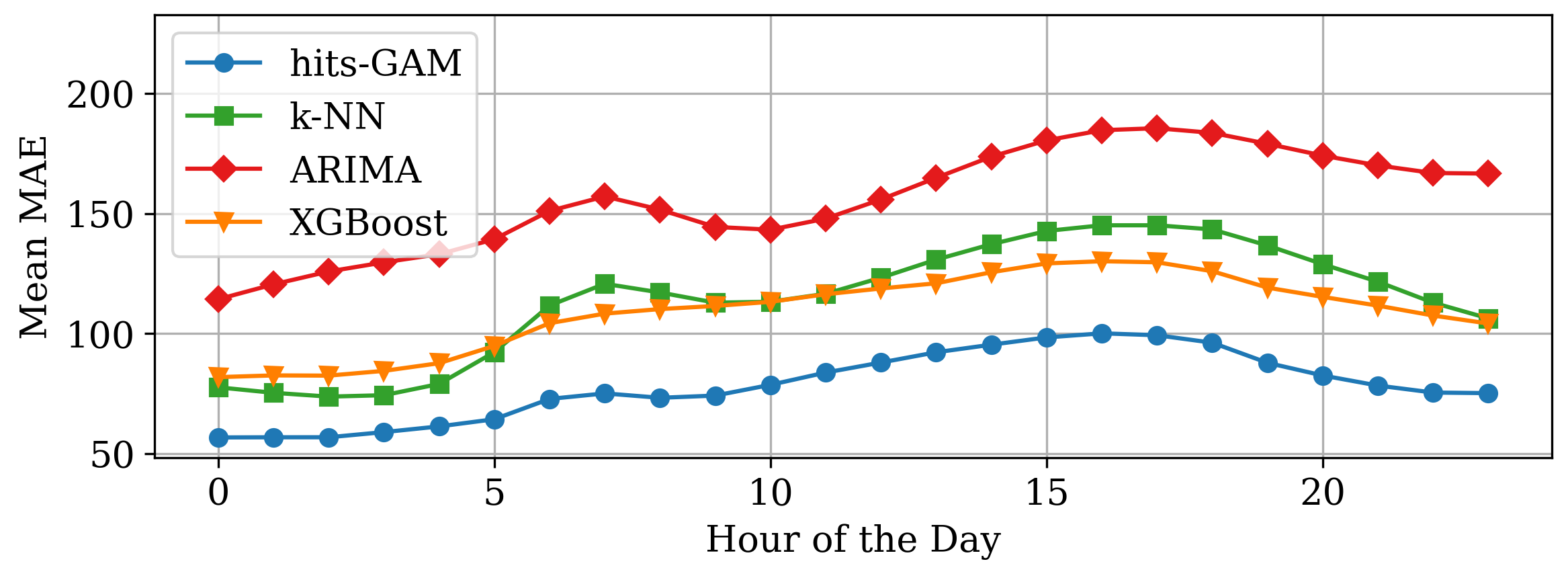} 
\caption{Average utility-level MAE [MW] by hour of day.}
\label{fig:mase_time}
\end{figure}

The hits-GAM model yields consistently accurate results across utilities, visualized in Figure \ref{fig:error_distribution}. Its lower median and a more compact interquartile range indicate superior accuracy and stability. 
Despite having been trained on scaled series, our model particularly improves accuracy for utilities with large errors. 
Metrics for individual utilities are presented in  Supplementary Material \ref{appendix-results}.
Our proposed model is also consistently the most accurate each hour of the day, illustrated in Figure \ref{fig:mase_time}. A common trend across all models is the error increasing as the forecast is projected further into the future, with a slight decline towards the end of the day due to stable night time demand.

\subsection{Model Interpretability}
\begin{figure*}[htb]
\includegraphics[width=\textwidth]{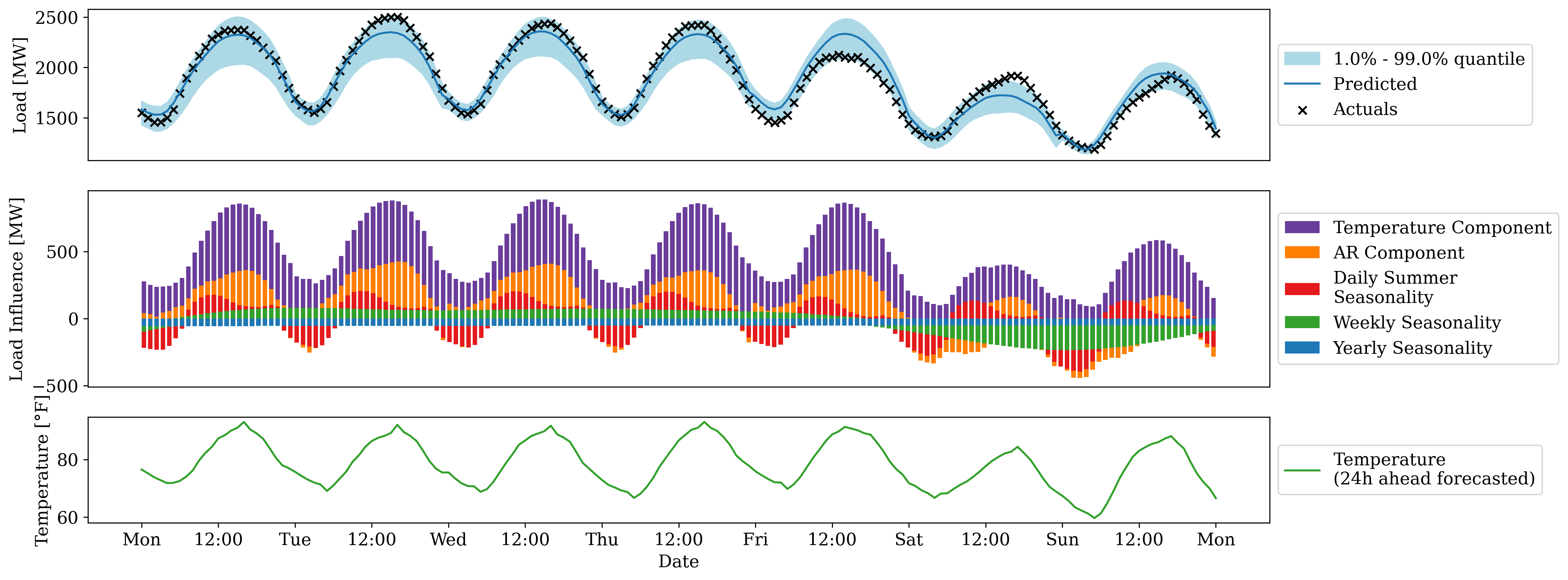}
\caption{
Day-ahead load forecasts: 
The upper plot shows the predicted and actual load, including the 98\% prediction interval. The middle plot delineates the contribution of each model component to the final forecast. 
The bottom plot shows the temperature forecast, ranging from 60 $^{\circ}F$ (15.6$^{\circ}C$) to 100 $^{\circ}F$ (37.8$^{\circ}C$), with averages around 80 $^{\circ}F$ (26.7$^{\circ}C$) .}
\label{fig:components}
\end{figure*}

We illustrate the forecast interpretability of hits-GAM by analyzing a week in June for an anonymous utility in Figure~\ref{fig:components}. 
The contributions of each model component are individually observable.
The model's temperature component substantially increases the predicted power demand during daytime hours when temperatures reach 100 $^{\circ}F$ (37.8$^{\circ}C$). 
From Saturday to Sunday temperatures drop, substantially reducing the forecast's temperature component during that period.
The Autoregressive (AR) component captures deviations of recent power demand. 
The lower than expected power demand of Friday is extrapolated by the AR component to continue on Saturday.
Concurrently, the forecast is affected by seasonality patterns, with daily seasonality higher during  day hours, weekly seasonality higher during weekdays, and yearly seasonality stable. Such a decomposition is also available for bus-level forecasts. 
We include a detailed analysis of the relationship of load to temperature in  Supplementary Material \ref{appendix-temp}.

\section{Bus Level Extension} \label{sec:case_study}  
Next, we evaluate the advantages and challenges of forecasting bus-level load and its integration with the utility level. In the following, we focus on Utility A and all its constituent buses. 
The bus level load time series are differently distributed and more heterogeneous than those at the utility level. The time series features described in Section \ref{sec:grouping-buses} exhibit a broader interquartile range at the bus level than at the utility level, and often their bus-level median falls outside the utility-level interquartile range.

Addressing these challenges, we evaluate three approaches of data pooling and hierarchical reconciliation using hits-GAM models with otherwise identical hyperparameters: 
\textit{`Local-Utility'}, a traditional local model for the utility-level time series, hierarchically reconciled to the bus level using a top-down approach;
\textit{`Global-Bus'}, a single global forecasting model fitted across all bus-level time series, hierarchically reconciled to the utility level using a bottom-up approach;
\textit{`Grouped Global-Bus'}, a variation of Global-Bus consisting of three models, each modeling a different subgroup of buses, as described in Section \ref{sec:grouping-buses}.

\begin{table}[!htbp]
\centering
\caption{Utility-Level Forecast Accuracy (Aggregated Bus Load)}
\label{tab:case-agg-bus}
\small
\begin{tabular}{lrrrr}
\hline
Model & RMSE & MAE & MASE & MSSE \\
\hline
Local-Utility  & 18.68 & 13.09 & 0.96 & 0.87 \\
Global-Bus & 18.70 & 12.75 & 0.72 & 0.62 \\
Grouped Global-Bus & \textbf{17.72} & \textbf{11.92} &  \textbf{0.68} & \textbf{0.55}\\
\hline
\end{tabular}
\end{table}

Table~\ref{tab:case-agg-bus} measures the accuracy of forecasting the aggregate load (sum of bus loads) for Utility A.
The Global-Bus approach performs similarly to the Local-Utility approach by absolute metrics, with higher accuracy on scaled metrics. 
The Grouped Global-Bus approach improves the forecast accuracy further, yielding the best results across all metrics.
An extension of this methodology, evaluated against the utility load series, is discussed in  Supplemental Material \ref{appendix-results}.

On a per-series basis, it is significantly harder to predict bus load due to data heterogeneity and volatility. 
This is reflected in an increase of MASE and MSSE in the bus-level Table~\ref{tab:case-bus-level}, compared to the utility-level Table~\ref{tab:case-agg-bus}. 
The degradation is particularly severe for the top-down Local-Utility approach, with scaled metrics well above one. The simple sNa\"ive prediction, by which the metrics are scaled, would yield MASE and MSSE scores of one.
However, both Global-Bus approaches are clearly more accurate than sNa\"ive.
This shows that disaggregating a utility-level forecast to the bus level is not accurate, whereas directly forecasting bus load is of strong predictive value.
More detailed statistics of the metrics are provided in  Supplementary Material \ref{appendix-results}. 

\begin{table}[!htbp]
\centering
\caption{Bus-Level Forecast Accuracy (Individual Bus Loads)}
\label{tab:case-bus-level}
\small
\begin{tabular}{lrrrr}
\hline
Model & RMSE & MAE & MASE & MSSE \\
\hline
Local-Utility  & 7.88 & 7.79 & 2.06 & 4.04 \\
Global-Bus & \textbf{0.58 }& \textbf{0.38} & \textbf{0.86} &\textbf{0.67} \\
Grouped Global-Bus & 0.59 & 0.39 & 0.87 & 0.74\\
\hline
\end{tabular}
\end{table}

Comparing the Grouped Global-Bus and Global-Bus approaches, we find grouping to improve accuracy across all metrics on the utility level in Table~\ref{tab:case-agg-bus}. 
This implies that the data heterogeneity issue of global models can be addressed by creating more homogeneous subgroups.
However, grouping slightly reduces accuracy on the bus level in Table~\ref{tab:case-bus-level}. Grouping reduces the number of series per global model, which increases the risk of overfitting due to bus-level volatility.

\subsection{Error Diagnosis}\label{error-diagnosis}
\begin{figure*}[hbt]
\centering
\includegraphics[width=\textwidth]{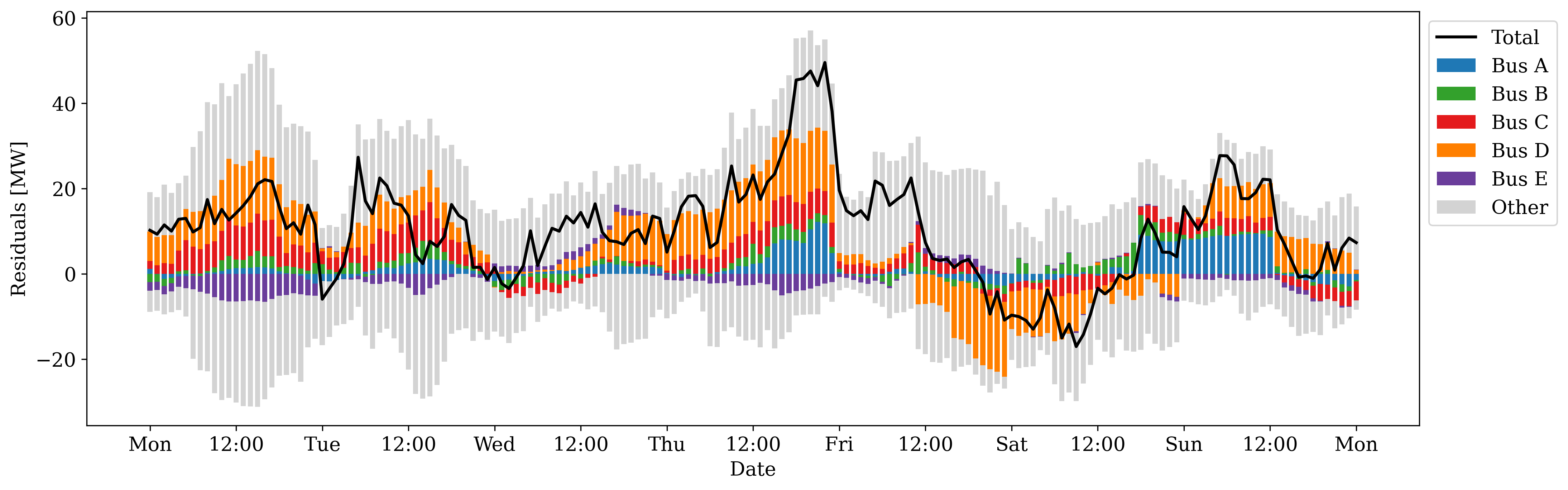}
\caption{
The Global-Bus utility-level error (black line) is given by the sum of individual bus forecast residuals (bars). The top five buses (colored bars) represent 45 percent of the utility's overall load. The remaining bus residuals are aggregated (grey bars).
}
\label{fig:case-bus-error-attribution}
\end{figure*}

Besides providing accurate forecasts, aggregating bus-level forecasts to the utility level allows us to attribute utility-level errors to individual buses.
Figure \ref{fig:case-bus-error-attribution} demonstrates the error-source attribution for an example week during summer.
On hours with high utility-level errors, most of the buses exhibit residuals in the same direction. For example, the co-occurrence of large errors in buses A and D drive such episodes on Thursday afternoon and Sunday before noon. 
However, low utility level errors do not always mean low bus level errors. On Friday, the utility error drops around noon, whereas the sum of bus errors increases. In particular, the residual of bus D flips in direction and increases in size, offsetting the residuals of other buses.

\subsubsection{Buses Driving Utility-Level Errors}
\begin{figure*}[hbt]
\centering
\includegraphics[width=\textwidth]{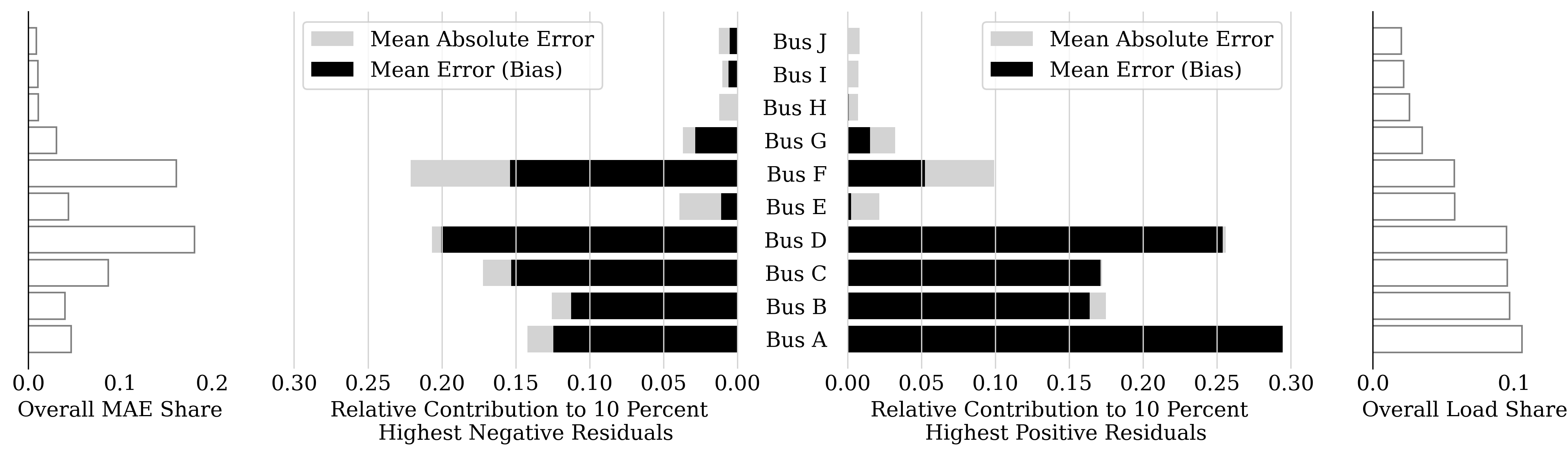}
\caption{
Correlation of individual bus errors to high-error occurrences at the utility level: 
Bus Mean Absolute Error (MAE) measures how hard a bus is to forecast (grey bar, including black bar), and Bus Mean Error (Bias) indicates how much a bus contributes to the overall error (black bar). 
If Bus Bias matches Bus MAE (black bar covers grey bar), all of its errors are directionally aligned with the utility level, exacerbating high-error situations. 
Bus Bias and Bus MAE are computed as mean error and mean absolute error, during high-error situations, both divided by the mean utility-level error.
Overall MAE share indicates how hard a bus is to forecast beyond high-error situations, relative to all buses.
Overall load share measures the relative average load magnitude of a bus. 
}
\label{fig:share_TopIds}
\end{figure*}

Situations pertaining to the 10~\% largest positive and negative utility-level residuals are analyzed in Figure \ref{fig:share_TopIds}.
Such an analysis is helpful to guide an operator's limited time and attention towards error-driving buses.
In high-error situations, the errors of buses A, B, C, and D are strong error drivers (high bias, covering MAE). Of these, A and B are more accurately forecastable (lower overall MAE share). Thus, high errors of buses A and B have a higher likelihood of directly contributing to high-error situations. 
While bus D strongly contributes to high-error situations, it is generally hard to forecast, and thus its errors do not necessarily contribute to a high-error situation. 
Buses F and bus A exhibit unique error dynamics, with their bias skewing in a direction, contributing predominantly to negative or positive high-error situations.
Based on this analysis, most attention should be directed towards buses A and B which possess strong error correlation, and later buses C, D and F that are strong error contributors, but less correlated. 
This information can also help a TSO's data scientists to prioritize and direct their model improvement efforts. 

\section{Discussion} \label{sec:discussion}  

We have demonstrated how our approach to multi-level forecasting using one interpretable model at the bus level can forecast load accurately at the utility level \textit{and} expose high spatial resolution information to TSO operators. Here, we discuss the practical implications for an operator. First, we highlight how the proposed forecasting system offers an operator a fine-grained way to connect forecasts and associated errors with real-world causes and effects. To illustrate this, we discuss practical examples of spatial nature in Section \ref{sec:discuss-spatial}, and temporal and spatio-temporal examples in Section \ref{sec:discuss-interpret}.  Second, we explore how our system can integrate with a TSO's existing operations and computational infrastructure in Section \ref{sec:discuss-integration}. Third, we discuss some of the limitations of our work and future opportunities in Section \ref{sec:discuss-limits}, with the preceding Section \ref{sec:discuss-tradeoff} detailing the implications of the volatility-heterogeneity trade-off.

\subsection{High Spatial Resolution}\label{sec:discuss-spatial}
Our interpretable system empowers TSO operators to adjust forecasts with confidence. Forecasts can be decomposed into temporal dynamics (model components) and segmented by spatial origin (bus forecasts). An operator trying to assess the reliability of a forecast would benefit substantially from having the information in Figures \ref{fig:components}, \ref{fig:case-bus-error-attribution}, and \ref{fig:share_TopIds} available. 
For example, in the days after a hurricane, the operator may wish to adjust the forecast down in order to account for disconnected buses. 
By looking at a plot which shows the contribution of each bus to the utility-level forecast -- like the middle panel of Figure \ref{fig:case-bus-error-attribution} but showing the actual forecast instead of forecast error -- they could identify how much each of the disconnected buses is contributing to the overall forecast. The regional (utility-level) forecast can then be adjusted down by the corresponding amount. This approach is applicable even if the operational forecast is produced by an independent forecasting system and not our proposed system.
Today such an adjustment needs to be done based on top-down forecast estimates of the fractional load historically represented by each bus, which is subject to high errors, as shown in Table \ref{tab:case-bus-level}. 

With our proposed forecasting system in operation, such procedures can even be automated -- which is key when dealing with a large number of buses. As our system's utility-level forecast stems from individual bus-level forecasts, operators can apply a situational load-factor from zero to one to all affected buses within a region. The factor can then be automatically applied to each bus' forecast, excluding disconnected buses, when they are summed up to the aggregate regional forecast.
Once the situation stabilizes and observations for the affected buses are available, an operator may choose to transition from factor-based corrections to analogous error-based corrections.
This allows for precise adjustments to the regional forecast without affecting the overall accuracy despite a system disruption.

\subsection{Forecast Interpretability}\label{sec:discuss-interpret}
Our proposed system meets the three TSO operator needs for interpretability outlined in Section \ref{interpretability-requirements}: The a) simple, intuitively understandable hits-GAM model with a b) clear input-output mapping given by each model component, is described in Section \ref{sec:proposed_model}, visualized in Figure \ref{fig:hitsGAM} and practically demonstrated in Section \ref{sec:results-utility}, especially in Figure \ref{fig:components}, including c) probabilistic forecasts. 
Using these tools, an operator can investigate how real-world load dynamics, forecasts and model errors are interconnected and address them with targeted and precise adjustments.
For example, at the beginning of the COVID-19 pandemic, the effects of lockdown policies were hard to model due to a lack of prior data. However, an operator might still be able to adjust forecasts more confidently using a model which is decomposable into components, such as those shown in Figure \ref{fig:components}.  As people stop commuting to the office, a TSO operator can expect each model component to be affected differently -- with the temperature's effect on load holding steady, and the weekly seasonality strongly affected. This enables targeted corrections of affected model components, without altering other components.

Furthermore, the increased spatial resolution allows for even more accurate diagnostics and adjustments of temporal changes to load dynamics. Expanding on the lockdown example above, buses that supply power to residential areas are expected to have higher load during weekdays, with load decreased for buses in commercial districts. By analyzing recent errors of individual buses, an operator can confirm their intuition and establish the quantity of adjustment needed per each individual bus or type of bus. Hereby, the operator may need to adjust the weekly seasonality of residential and commercial buses in opposite directions, making bus-level adjustments significantly more effective than applying an average adjustment across all buses. This illustrates how the ability to disaggregate forecasts into components \textit{and} individual buses allows an operator to identify specific load dynamics changes and make precise forecast adjustments.

In this section and the previous one, we gave spatial, temporal, and spatio-temporal examples of system perturbations, which demonstrate how forecast decomposition can directly help TSO operators make better forecast adjustments for real-time operations. Bus forecasts also reveal bus error-cancellation effects, signaling implicit uncertainty which is relevant for an operator's risk assessment.
When refining the model, a TSO's data scientists can use analogous decompositions of error to prioritize their work, as discussed in Section \ref{error-diagnosis} and visualized in Figures \ref{fig:case-bus-error-attribution} and \ref{fig:share_TopIds}. 

\subsection{System Manageability and Operational Integration}\label{sec:discuss-integration}
The proposed system \textit{not only} allows TSO operators to keep their current workflow of monitoring the utility-level forecasts, \textit{but also} gives them the option to drill down into the details of specific bus-level forecasts and specific model components.
A TSO seeking to deploy this system would integrate it as a component downstream of their existing energy management system (EMS). For a TSO already collecting high-frequency bus-level load data, like MISO, no new data collection and control systems would be needed.
Initially, the system could be run in parallel to the current system, which would already offer additional insights to operators. However, a full transition does not require large operational changes as the utility-level forecasts remain center-stage for operators, while long-term benefits can be observed from precise forecast adjustments and model refinement.

In terms of computational cost, the model inference cost would grow approximately linearly with the number of buses in the system. This is because the global model structure requires one forecast inference for each bus in the system. However, there are no interactions between buses which would cause a super-linear increase in complexity with system size. On the contrary, our global model implementation is fully parallelized for each forecasted bus and supports GPU-acceleration. While scaling from one to one hundred buses and beyond increases training complexity in a sub-linear manner, training time can be reduced with the `Grouped Global Bus' approach, as each of the models is trained on a subset of data, which allows for parallel computation on independent systems. This should not pose a challenge even with thousands of buses, as retraining is expected to be executed once every few weeks to days. Notably, inference time is essentially independent of the number of buses due to the full parallelization of our forecasting system. This makes it possible to forecast all buses of an entire TSO system in one shot with modern computational infrastructure. 

Indeed, in our experience we find that training of our hits-GAM model for all the buses considered in this study takes less than an hour, while an inference run producing a full hourly day-ahead forecast for all buses takes less than a second on  a single GPU.
In our application, the forecast is expected to be computed once a day to support day-ahead operations, as outlined in Section \ref{sec:experiments}. However, thanks to the inference speed, the presented forecasting system is also suitable for applications requiring more frequent updates, such as ultra-short-term forecasts generated every minute.

\subsection{Volatility-Heterogeneity Trade-off}\label{sec:discuss-tradeoff}
Clustering by time series characteristics reduces intra-group heterogeneity and increases inter-group heterogeneity, resulting in a unique model fit per group. Akin to ensembling~\cite{Lu2022ModelModel}, this model diversity can improve aggregate forecast accuracy -- despite some overfitting risk in individual models -- by reduced correlation of errors across groups. 
Our results on the effects of grouping show promise to address heterogeneity, but their increased susceptibility to volatility highlights that GFMs are subject to a volatility-heterogeneity tradeoff. This motivates future research to investigate the effects of grouping mechanisms and group sizes on GFMs.

\subsection{Limitations and Future Work}\label{sec:discuss-limits}
Considering the limitations of our underlying model, we acknowledge that the strictly additive structure of hits-GAM provides interpretability benefits, but prevents interaction effects between components. 
For example, the weekly seasonality effect is modeled as a static offset. This might not be sufficient if the temperature-load relationship is different on weekdays than on weekends. Addressing such dynamic interaction effects requires either a more complicated model expressly modeling specific conditional situations or a more complex model that flexibly models all effects in a black box. Both of these options trade interpretability for accuracy to a point where the outlined interpretability requirements of TSO operators are violated. Novel approaches to model component interaction effects addressing the accuracy-interpretability trade-off could be the subject of future work.
Hereby we note that hits-GAM incorporates uncertainty estimates unlike other benchmarked methods. The joint fitting of multiple objectives, i.e. mean estimates alongside uncertainty, generally leads to a trade-off in accuracy, but benefits interpretation due to coherence. 

A promising direction for future work is to develop a methodology to directly quantify the uncertainty implied by bus error-cancellation effects. Such a methodology would benefit an operator's risk assessment efforts but require advanced and novel hierarchical reconciliation methods to retain the forecasting system's interpretability. 
A further avenue for future research includes an investigation into spatial patterns, requiring the availability and integration of geospatial data. 
Another limitation of our study is the use of static graphs, such as Figures \ref{fig:components} and \ref{fig:case-bus-error-attribution}, to visualize the benefits of the proposed system. In the future, an interactive interface could be built for TSO operators which allows them to drill down into the forecast details when necessary. Such a user interface could support the dynamic navigation between graphs and levels  -- another potential direction for future research.

\section{Conclusion} \label{sec:conclusion}
Extending TSO utility-level forecasting operations with coherent bus-level forecasts is key, as important information is otherwise lost in aggregation. 
Our proposed multi-level forecasting system is expected to improve forecast accuracy and to empower operators in their key efforts.
Using our uniquely extensive, multi-level TSO dataset, our results show that:

\begin{enumerate}[I.,topsep=4pt,parsep=0pt,leftmargin=24pt]
\item Our hits-GAM model outperforms established models when trained at the utility level alone, with a 24\% reduction in MAPE compared to XGBoost.
In addition, hits-GAM exhibits strong interpretability benefits, such as forecast decomposition and uncertainty estimates, scales to predict many independent series at once, and generalizes to series of varying magnitudes from small nodes to large zonal aggregates.
\item Our multi-level system, trained on bus-level data alone, further reduces the utility-level error by 5\% (RMSE) and 9\% (MAE) compared to a local model. 
\item Our system improves bus-level accuracy substantially, reducing the error by 33\% (MSSE) and 14\% (MASE) compared to sNa\"ive, and 93\% (RMSE) and 95\% (MAE) compared to a top-down disaggregation. 
\end{enumerate}

Hereby, our system is composed of an interpretable yet scalable model and a grouped global training paradigm with bottom-up hierarchical reconciliation.
Based on an analysis of the results, we derive the following implications of our approach:

\begin{enumerate}[A.,topsep=4pt,parsep=0pt,leftmargin=24pt]
\item The volatile and heterogeneous nature of bus loads is partially mitigated by our system using a global training paradigm with subgrouping, subject to a trade-off. 
\item Our system is highly manageable thanks to an interpretable single-model workflow that is fully parallelized and independent of the number of buses, with sub-second inference. 
\item In practice, our proposed multi-level forecasting system enables operators to make precise forecast adjustments, understand specific load dynamics, accurately attribute error sources, manage risks, diagnose and improve their model. 
\end{enumerate}

The integration of high-resolution forecasts is a challenging, yet beneficial research area, essential to the reliable operation of a sustainable power grid.
We demonstrate a multi-level forecasting system which achieves high spatial resolution, accuracy, interpretability, and manageability. 
We hope to encourage more work from the perspective of TSO employee-operators -- the guardians of reliability in the power grid. 

\section*{Acknowledgements, Disclaimers and Declarations}
We are grateful for the collaboration with MISO, including their support with data and their domain expertise. In particular, we would like to thank Michael Parran, Carmen Pippenger, Congcong Wang, Adam Simkowski, Long Zhao, Amir Quadri, Bonnie Mathew, Craig Wilkert, and Adela Creasy.

The claims, views, and opinions expressed in this paper are solely those of the authors and do not necessarily represent any policy or position of Midcontinent Independent System Operator, Inc.

The authors declare the following financial interests/personal relationships which may be considered as potential competing interests:
Ram Rajagopal has received a gift from MISO to pursue research of his choosing at Stanford. 
MISO has further provided Stanford with access to computational resources on Azure.
Arezou Ghesmati and Chen-Hao Tsai are employed by MISO.
Christoph Bergmeir is supported by a María Zambrano Fellowship that is funded by the Spanish Ministry of Universities and Next Generation funds from the European Union.
The other authors declare that they have no known competing financial interests or personal relationships that could have appeared to influence the work reported in this paper.

During the preparation of this work the authors used ChatGPT in order to improve the readability and language of the manuscript. After using this tool, the authors reviewed and edited the content as needed and take full responsibility for the content of the published article.

\singlespacing
\bibliographystyle{elsarticle-num-names} 
\bibliography{ref_all_arxiv}

\begin{thebibliography}{38}
\expandafter\ifx\csname natexlab\endcsname\relax\def\natexlab#1{#1}\fi
\providecommand{\url}[1]{\texttt{#1}}
\providecommand{\href}[2]{#2}
\providecommand{\path}[1]{#1}
\providecommand{\DOIprefix}{doi:}
\providecommand{\ArXivprefix}{arXiv:}
\providecommand{\URLprefix}{URL: }
\providecommand{\Pubmedprefix}{pmid:}
\providecommand{\doi}[1]{\href{http://dx.doi.org/#1}{\path{#1}}}
\providecommand{\Pubmed}[1]{\href{pmid:#1}{\path{#1}}}
\providecommand{\bibinfo}[2]{#2}
\ifx\xfnm\relax \def\xfnm[#1]{\unskip,\space#1}\fi
\bibitem[{Gielen et~al.(2019)}]{gielen_role_2019}
\bibinfo{author}{D.~Gielen}, et~al.,
\newblock \bibinfo{title}{The role of renewable energy in the global energy transformation},
\newblock \bibinfo{journal}{Energy strategy reviews} \bibinfo{volume}{24} (\bibinfo{year}{2019}) \bibinfo{pages}{38--50}.
\bibitem[{Haben et~al.(2021)}]{lv-review}
\bibinfo{author}{S.~Haben}, et~al.,
\newblock \bibinfo{title}{Review of low voltage load forecasting: Methods, applications, and recommendations},
\newblock \bibinfo{journal}{Applied Energy} \bibinfo{volume}{304} (\bibinfo{year}{2021}) \bibinfo{pages}{117798}.
\bibitem[{Wang et~al.(2023)}]{wang_geospatial_2023}
\bibinfo{author}{Z.~Wang}, et~al.,
\newblock \bibinfo{title}{Geospatial mapping of distribution grid with machine learning and publicly-accessible multi-modal data},
\newblock \bibinfo{journal}{Nature Communications} \bibinfo{volume}{14} (\bibinfo{year}{2023}) \bibinfo{pages}{5006}.
\bibitem[{Schröter et~al.(2020)}]{substation-storage-5-outof-56}
\bibinfo{author}{T.~Schröter}, et~al.,
\newblock \bibinfo{title}{Substation related forecasts of electrical energy storage systems: Transmission system operator requirements},
\newblock \bibinfo{journal}{Energies} \bibinfo{volume}{13} (\bibinfo{year}{2020}).
\bibitem[{Sun et~al.(2013)}]{sun2013efficient}
\bibinfo{author}{X.~Sun}, et~al.,
\newblock \bibinfo{title}{An efficient approach for short-term substation load forecasting},
\newblock in: \bibinfo{booktitle}{2013 IEEE Power \& Energy Society General Meeting}, \bibinfo{organization}{IEEE}, \bibinfo{year}{2013}, pp. \bibinfo{pages}{1--5}.
\bibitem[{Wang et~al.(2019)}]{tao-hong-smart-meter-agg}
\bibinfo{author}{Y.~Wang}, et~al.,
\newblock \bibinfo{title}{Review of smart meter data analytics: Applications, methodologies, and challenges},
\newblock \bibinfo{journal}{IEEE Transactions on Smart Grid} \bibinfo{volume}{10} (\bibinfo{year}{2019}) \bibinfo{pages}{3125--3148}.
\bibitem[{Sevlian and Rajagopal(2018)}]{ram-scaling-law}
\bibinfo{author}{R.~Sevlian}, \bibinfo{author}{R.~Rajagopal},
\newblock \bibinfo{title}{A scaling law for short term load forecasting on varying levels of aggregation},
\newblock \bibinfo{journal}{International Journal of Electrical Power \& Energy Systems} \bibinfo{volume}{98} (\bibinfo{year}{2018}) \bibinfo{pages}{350--361}.
\bibitem[{Abdolrezaei et~al.(2022)}]{abdolrezaei2022substation}
\bibinfo{author}{H.~Abdolrezaei}, et~al.,
\newblock \bibinfo{title}{Substation mid-term electric load forecasting by knowledge-based method},
\newblock \bibinfo{journal}{Energy, Ecology and Environment}  (\bibinfo{year}{2022}) \bibinfo{pages}{1--11}.
\bibitem[{Chen et~al.(2025)}]{CHEN2025epsr}
\bibinfo{author}{Y.~Chen}, et~al.,
\newblock \bibinfo{title}{Day-ahead bus load forecasting method based on fully connected spatial-temporal graph attention network},
\newblock \bibinfo{journal}{Electric Power Systems Research} \bibinfo{volume}{241} (\bibinfo{year}{2025}) \bibinfo{pages}{111294}.
\bibitem[{Su et~al.(2024)}]{tampa-1-campus}
\bibinfo{author}{Z.~Su}, et~al.,
\newblock \bibinfo{title}{Short-term load forecasting of regional integrated energy system based on spatio-temporal convolutional graph neural network},
\newblock \bibinfo{journal}{Electric Power Systems Research} \bibinfo{volume}{232} (\bibinfo{year}{2024}) \bibinfo{pages}{110427}.
\bibitem[{Lizhen et~al.(2022)}]{map-reduce-2-series}
\bibinfo{author}{W.~Lizhen}, et~al.,
\newblock \bibinfo{title}{A novel short-term load forecasting method based on mini-batch stochastic gradient descent regression model},
\newblock \bibinfo{journal}{Electric Power Systems Research} \bibinfo{volume}{211} (\bibinfo{year}{2022}) \bibinfo{pages}{108226}.
\bibitem[{He et~al.(2025)}]{gru-stlf-2-areas}
\bibinfo{author}{X.~He}, et~al.,
\newblock \bibinfo{title}{Short-term load forecasting by gru neural network and ddpg algorithm for adaptive optimization of hyperparameters},
\newblock \bibinfo{journal}{Electric Power Systems Research} \bibinfo{volume}{238} (\bibinfo{year}{2025}) \bibinfo{pages}{111119}.
\bibitem[{Chen et~al.(1996)}]{1996-ann-3-sub}
\bibinfo{author}{C.~Chen}, et~al.,
\newblock \bibinfo{title}{The application of artificial neural networks to substation load forecasting},
\newblock \bibinfo{journal}{Electric Power Systems Research} \bibinfo{volume}{38} (\bibinfo{year}{1996}) \bibinfo{pages}{153--160}.
\bibitem[{Nose-Filho et~al.(2011)}]{nine-substations}
\bibinfo{author}{K.~Nose-Filho}, et~al.,
\newblock \bibinfo{title}{Short-term multinodal load forecasting using a modified general regression neural network},
\newblock \bibinfo{journal}{IEEE Transactions on Power Delivery} \bibinfo{volume}{26} (\bibinfo{year}{2011}) \bibinfo{pages}{2862--2869}.
\bibitem[{Ferreira et~al.(2025)}]{transformer-nz}
\bibinfo{author}{A.~Ferreira}, et~al.,
\newblock \bibinfo{title}{Power substation load forecasting using interpretable transformer-based temporal fusion neural networks},
\newblock \bibinfo{journal}{Electric Power Systems Research} \bibinfo{volume}{238} (\bibinfo{year}{2025}) \bibinfo{pages}{111169}.
\bibitem[{Evangelopoulos and Georgilakis(2022)}]{ltlf-spatial}
\bibinfo{author}{V.~A. Evangelopoulos}, \bibinfo{author}{P.~S. Georgilakis},
\newblock \bibinfo{title}{Probabilistic spatial load forecasting for assessing the impact of electric load growth in power distribution networks},
\newblock \bibinfo{journal}{Electric Power Systems Research} \bibinfo{volume}{207} (\bibinfo{year}{2022}) \bibinfo{pages}{107847}.
\bibitem[{Mathew et~al.(2024)}]{mtlf-feeder}
\bibinfo{author}{A.~Mathew}, et~al.,
\newblock \bibinfo{title}{Medium-term feeder load forecasting and boosting peak accuracy prediction using the pwp-xgboost model},
\newblock \bibinfo{journal}{Electric Power Systems Research} \bibinfo{volume}{237} (\bibinfo{year}{2024}) \bibinfo{pages}{111051}.
\bibitem[{Chen et~al.(2023)}]{chen_interpretable_2023}
\bibinfo{author}{Z.~Chen}, et~al.,
\newblock \bibinfo{title}{Interpretable machine learning for building energy management: {A} state-of-the-art review},
\newblock \bibinfo{journal}{Advances in Applied Energy}  (\bibinfo{year}{2023}) \bibinfo{pages}{100123}.
\bibitem[{Pinheiro et~al.(2023)}]{pinheiro2023short}
\bibinfo{author}{M.~G. Pinheiro}, et~al.,
\newblock \bibinfo{title}{Short-term electricity load forecasting -- a systematic approach from system level to secondary substations},
\newblock \bibinfo{journal}{Applied Energy} \bibinfo{volume}{332} (\bibinfo{year}{2023}) \bibinfo{pages}{120493}.
\bibitem[{Hastie and Tibshirani(2017)}]{Hastie2017GeneralizedModels}
\bibinfo{author}{T.~Hastie}, \bibinfo{author}{R.~Tibshirani}, \bibinfo{title}{{Generalized Additive Models}}, \bibinfo{publisher}{Routledge}, \bibinfo{year}{2017}.
\bibitem[{Hyndman and Athanasopoulos(2024)}]{hyndman2018}
\bibinfo{author}{R.~J. Hyndman}, \bibinfo{author}{G.~Athanasopoulos}, \bibinfo{title}{{Forecasting: principles and practice, 3rd edition}}, \bibinfo{publisher}{OTexts}, \bibinfo{year}{Accessed on June 2024}.
\bibitem[{Hewamalage et~al.(2022)}]{hemawalage-gfm}
\bibinfo{author}{H.~Hewamalage}, et~al.,
\newblock \bibinfo{title}{Global models for time series forecasting: A simulation study},
\newblock \bibinfo{journal}{Pattern Recognition} \bibinfo{volume}{124} (\bibinfo{year}{2022}) \bibinfo{pages}{108441}.
\bibitem[{Hyndman et~al.(2020)}]{Hyndman2020Time1.0.2}
\bibinfo{author}{R.~J. Hyndman}, et~al.,
\newblock \bibinfo{title}{{Time Series Feature Extraction}},
\newblock in: \bibinfo{booktitle}{R package tsfeatures [version 1.0.2]}, \bibinfo{year}{2020}.
\bibitem[{Triebe et~al.(2019)}]{Triebe2019AR-Net:Time-series}
\bibinfo{author}{O.~Triebe}, et~al.,
\newblock \bibinfo{title}{{AR-Net: A simple Auto-Regressive Neural Network for time-series}},
\newblock \bibinfo{journal}{arXiv}  (\bibinfo{year}{2019}).
\bibitem[{Koenker and Bassett(1978)}]{Koenker1978RegressionQuantiles}
\bibinfo{author}{R.~Koenker}, \bibinfo{author}{G.~Bassett},
\newblock \bibinfo{title}{{Regression Quantiles}},
\newblock \bibinfo{journal}{Econometrica} \bibinfo{volume}{46} (\bibinfo{year}{1978}) \bibinfo{pages}{33}.
\bibitem[{Steinwart and Christmann(2011)}]{Steinwart2011EstimatingLoss}
\bibinfo{author}{I.~Steinwart}, \bibinfo{author}{A.~Christmann},
\newblock \bibinfo{title}{{Estimating conditional quantiles with the help of the pinball loss}},
\newblock \bibinfo{journal}{Bernoulli} \bibinfo{volume}{17} (\bibinfo{year}{2011}).
\bibitem[{Triebe et~al.(2021)}]{Triebe2021NeuralProphet:Scale}
\bibinfo{author}{O.~Triebe}, et~al.,
\newblock \bibinfo{title}{{NeuralProphet: Explainable Forecasting at Scale}},
\newblock \bibinfo{journal}{arXiv}  (\bibinfo{year}{2021}).
\bibitem[{Stromer et~al.(2023)}]{Stromer2023DesigningForecasts}
\bibinfo{author}{R.~Stromer}, et~al.,
\newblock \bibinfo{title}{{Designing forecasting software for forecast users: Empowering non-experts to create and understand their own forecasts}},
\newblock in: \bibinfo{booktitle}{Americas Conference on Information Systems}, \bibinfo{year}{2023}.
\bibitem[{Valgaev et~al.(2016)}]{knn-load}
\bibinfo{author}{O.~Valgaev}, et~al.,
\newblock \bibinfo{title}{Low-voltage power demand forecasting using k-nearest neighbors approach},
\newblock in: \bibinfo{booktitle}{2016 IEEE Innovative Smart Grid Technologies - Asia (ISGT-Asia)}, \bibinfo{year}{2016}, pp. \bibinfo{pages}{1019--1024}.
\bibitem[{Chen and Guestrin(2016)}]{Chen2016XGBoost:System}
\bibinfo{author}{T.~Chen}, \bibinfo{author}{C.~Guestrin},
\newblock \bibinfo{title}{{XGBoost: A Scalable Tree Boosting System}},
\newblock \bibinfo{journal}{Proceedings of the ACM SIGKDD International Conference on Knowledge Discovery and Data Mining} \bibinfo{volume}{13-17-August-2016} (\bibinfo{year}{2016}) \bibinfo{pages}{785--794}.
\bibitem[{Makridakis et~al.(2020)}]{Makridakis2020}
\bibinfo{author}{S.~Makridakis}, et~al.,
\newblock \bibinfo{title}{{The M4 Competition: 100,000 time series and 61 forecasting methods}},
\newblock \bibinfo{journal}{International Journal of Forecasting} \bibinfo{volume}{36} (\bibinfo{year}{2020}) \bibinfo{pages}{54--74}.
\bibitem[{Rostami-Tabar and Hyndman(2024)}]{Rostami-TabaraHierarchicalServices}
\bibinfo{author}{B.~Rostami-Tabar}, \bibinfo{author}{R.~J. Hyndman},
\newblock \bibinfo{title}{Hierarchical time series forecasting in emergency medical services},
\newblock \bibinfo{journal}{Journal of Service Research} \bibinfo{volume}{0} (\bibinfo{year}{2024}) \bibinfo{pages}{10946705241232169}.
\bibitem[{Lu and Shi(2022)}]{Lu2022ModelModel}
\bibinfo{author}{W.~Lu}, \bibinfo{author}{W.~Shi},
\newblock \bibinfo{title}{{Model Averaging Estimation Method by Kullback–Leibler Divergence for Multiplicative Error Model}},
\newblock \bibinfo{journal}{Complexity} \bibinfo{volume}{2022} (\bibinfo{year}{2022}) \bibinfo{pages}{1--13}.
\bibitem[{Chowdhury and Islam(2021)}]{Chowdhury2021ARIMACSE}
\bibinfo{author}{T.~U. Chowdhury}, \bibinfo{author}{M.~S. Islam},
\newblock \bibinfo{title}{{ARIMA Time Series Analysis in Forecasting Daily Stock Price of Chittagong Stock Exchange (CSE)}},
\newblock \bibinfo{journal}{International Journal of Research and Innovation in Social Science} \bibinfo{volume}{05} (\bibinfo{year}{2021}) \bibinfo{pages}{214--233}.
\bibitem[{Herzen et~al.(2022)}]{Herzen2022Darts:Series}
\bibinfo{author}{J.~Herzen}, et~al.,
\newblock \bibinfo{title}{{Darts: User-Friendly Modern Machine Learning for Time Series}},
\newblock \bibinfo{journal}{Journal of Machine Learning Research} \bibinfo{volume}{23} (\bibinfo{year}{2022}) \bibinfo{pages}{1--6}.
\bibitem[{Grinsztajn et~al.(2022)}]{Grinsztajn2022WhyData}
\bibinfo{author}{L.~Grinsztajn}, et~al.,
\newblock \bibinfo{title}{{Why do tree-based models still outperform deep learning on tabular data?}}  (\bibinfo{year}{2022}).
\bibitem[{Wang and Ni(2019)}]{Wang2019AOptimization}
\bibinfo{author}{Y.~Wang}, \bibinfo{author}{X.~S. Ni},
\newblock \bibinfo{title}{{A XGBoost risk model via feature selection and Bayesian hyper-parameter optimization}},
\newblock \bibinfo{journal}{International Journal of Database Management Systems} \bibinfo{volume}{11} (\bibinfo{year}{2019}) \bibinfo{pages}{01--17}.
\bibitem[{Alanazi et~al.(2016)}]{alanazi_long-term_2016}
\bibinfo{author}{M.~Alanazi}, et~al.,
\newblock \bibinfo{title}{Long-term solar generation forecasting},
\newblock in: \bibinfo{booktitle}{2016 {IEEE}/{PES} transmission and distribution conference and exposition ({T}\&{D})}, \bibinfo{publisher}{IEEE}, \bibinfo{year}{2016}, pp. \bibinfo{pages}{1--5}.

\end{thebibliography}



\appendix
\clearpage
\section{Supplementary Material on Methods}\label{appendix-methods}
Supplementary material for manuscript 
"Extending Load Forecasting from Zonal Aggregates to Individual Nodes for Transmission System Operators``
submitted to ``Sustainable Energy, Grids and Networks''
by Oskar Triebe et al.  (2025).

\subsection{Datasets}\label{appendix:data_description}
This study employs an extensive dataset subdivided into two aggregation levels: utility and bus.
Due to differences in measurement methods, the two dataset levels do not match perfectly when aggregated, despite representing the same power demand. 

\begin{figure}[htbp]
\centering
\includegraphics[width=0.55\textwidth]{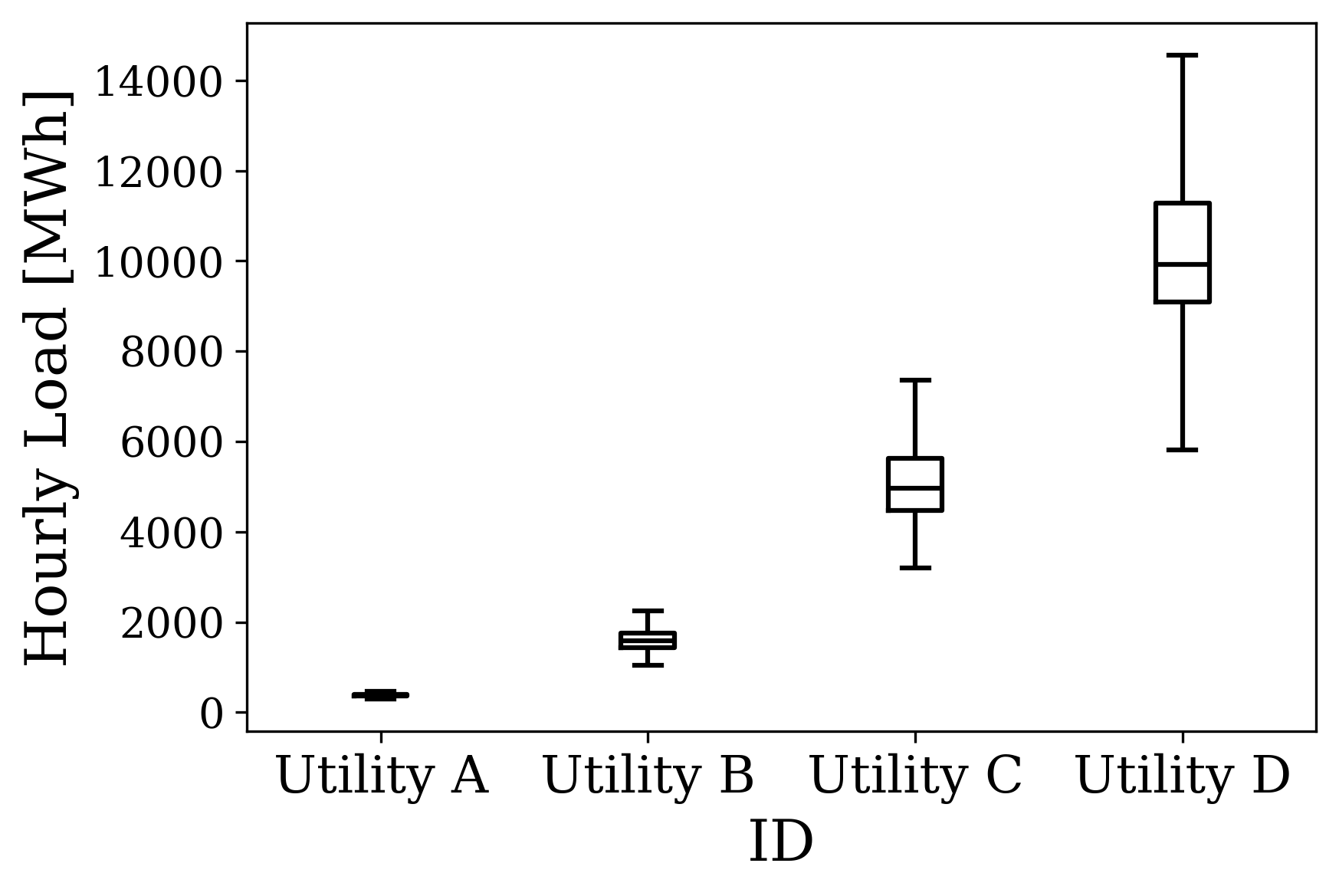}
\caption{Distribution of hourly aggregate load for select utilities.
}
\label{fig:load_distribution}
\end{figure}

\subsubsection{Utility Dataset}
The utility dataset features hourly net energy demand in megawatts (MW) for 37 aggregation zones (utilities) at the high-voltage transmission level.
The mean hourly load for each utility ranges from 102 MWh to 10,300 MWh.
The temporal coverage of the dataset spans from January 2018 to September 2021. 
Data preprocessing, including duplicate removal and linear imputation of missing values, led to negligible differences in total load post-cleaning.
For selected utilities load distribution is shown in Figure \ref{fig:load_distribution}, representative of the dataset range. 

\subsubsection{Case Study Bus Dataset}\label{sec:bus_data_characteristics}
A case study is conducted on the subset of the buses within Utility A. 
Utility A is one of the smaller aggregation zones with average hourly bus loads ranging from 0.16~MWh to 37~MWh. 
The bus data preprocessing pipeline includes duplicate removal, outlier replacement, and missing value handling. 
Data points three standard deviations above the mean are considered outliers and are replaced using linear interpolation. 
Hereby, missing values are imputed by linear interpolation for up to 20 consecutive missing entries, beyond which a rolling average imputation is applied. 
Buses with less than one year of training data, more than 20\% missing values, only constant values, or missing more than 15 days at the end are removed entirely. 
After data cleaning, the number of buses was reduced by approximately one quarter to about 100 buses.
After preprocessing, the cleaned bus dataset deviates from the raw bus dataset by 2.89\% in aggregate, measured as hourly Mean Absolute Percentage Error (MAPE).

\subsection{Model Configurations}

\subsubsection{hits-GAM}
Table \ref{tab:neuralprophet_params} describes the chosen hyperparameters for hits-GAM.
The autoregressive neural network (AR-Net) components takes observations of the last 15 days as input, the seasonal components are modeled with Fourier terms with periodicities representing yearly, weekly, and daily seasonality. The daily seasonality is conditional, i.e. a separate seasonal component is fitted for summer (April to September) and winter (October to March). 

\begin{table}[htbp]
\centering
\caption{hits-GAM Parameters}
\label{tab:neuralprophet_params}
\small
\begin{tabular}{ll}
\hline
Parameter & Value \\
\hline
n\_lags & \(24 \times 15\) \\
newer\_samples\_weight & 2.0 \\
n\_changepoints & 0 \\
yearly\_seasonality & 10 \\
weekly\_seasonality & True \\ 
daily\_seasonality & True (conditional, summer and winter) \\
batch\_size & 128 \\
ar\_layers & [32, 64, 32, 16] \\
lagged\_reg\_layers & [32, 32] \\
learning\_rate & 0.001 \\
epochs & 30 \\
trend\_global\_local & local \\
season\_global\_local & local \\
normalize & standardize \\
\hline
\end{tabular}
\end{table}

\subsubsection{Seasonal ARIMA}
ARIMA~\cite{hyndman2018} is included as benchmark model due to its robust statistical underpinnings and its ubiquity in the realm of univariate time series forecasting. Notably, the model is interpretable, with the parameters p and q showing how many recent observations and errors contribute to each forecast step in the horizon, respectively. The fitted coefficients allow to directly relate a specific observation's contribution to a specific forecast target \cite{hyndman2018}.
A seasonal variant of ARIMA is employed, with parameter selection guided by the Augmented Dickey-Fuller test, AutoCorrelation Function (ACF), and Partial AutoCorrelation Function (PACF) plots, as delineated by Chowdhury et al. \cite{Chowdhury2021ARIMACSE}. The model is instantiated using the Darts library \cite{Herzen2022Darts:Series}, yielding the parameters \( p=1 \), \( d=0 \), \( q=1 \), and \texttt{seasonal\_order}=\((1, 0, 1, 24)\). The seasonal terms are multiplied by the non-seasonal terms, effectively considering up to 24 lagged values. In a previous attempt, we tried implementing AutoARIMA using the Darts library. However, due to the excessively long computation time, automatic parameter selection was not feasible.

\subsubsection{XGBoost}
The XGBoost model~\cite{Chen2016XGBoost:System} is incorporated into the evaluation framework due to its consistent performance across a multitude of machine learning competitions \cite{Grinsztajn2022WhyData}. The model's configuration is harmonized with that of the proposed hits-GAM, employing identical historical data and temperature forecasts as features. Additionally, hyperparameter tuning is conducted for XGBoost in accordance with established methodologies \cite{Wang2019AOptimization}, and the model is implemented using the Darts library \cite{Herzen2022Darts:Series}.

\subsubsection{kNN}
The kNN algorithm~\cite{knn-load} is chosen for its consistent use as load forecasting baseline and inherent interpretability. The model utilizes historical load data to inform its forecasts, rendering it an intuitively appealing choice for this study. The feature vector comprises both temperature forecasts for the day under prediction and the preceding 24-hour load data. The hyperparameter \( k \), representing the number of nearest neighbors, is empirically set to 3.

\subsubsection{sNa\"ive}
The seasonal naive forecast (sNa\"ive)~\cite{hyndman2018} serves as the most straightforward yet effective baseline, including as the as the base forecast of the scaled metrics MASE and MSSE. The sNa\"ive model simply uses the most recent observations, in the window of 24 to 48 hours prior, as its forecast. This 48-hour lag is chosen because the data from 24 hours prior is not fully available at the time of forecast generation, at 2pm of the compute day.

\subsubsection{Model Training Infrastructure} \label{appendix:hyperparams}
The computational workload is executed on a A100 Virtual Machine, outfitted with an NVIDIA A100 PCIe GPU and 3rd-generation AMD EPYC\textsuperscript{TM} 7V13 (Milan) processors. The machine is configured with 24 CPU cores, 220 GiB of RAM, a single GPU, and 80 GiB of GPU memory.
Hyperparameters are set according to domain-typical ranges and kept identical across comparison models where possible (e.g. 15 days of lagged observations).

\subsection{Forecasting Methods}

\subsubsection{Forecast Time Horizon}
The forecast time horizon varies, ranging from seconds to decades, with corresponding variability in suitable applications. Commonly used time horizons and common application examples include \cite{pinheiro2023short, alanazi_long-term_2016}:
\paragraph{Very short-term} Seconds to hours,  e.g. an hour-ahead forecast with five-minute intervals for operational corrective adjustments, real-time price.
\paragraph{Short-term} Hours to weeks, e.g. hourly day-ahead forecast for prediction of demand and generation scheduling, operation, maintenance and trading.
\paragraph{Medium-term} Weeks to years, e.g. daily year-ahead forecast for generation and consumption trends, project queue management, logistics and trading.
\paragraph{Long-term} Years to decades, e.g. quarterly decade-ahead forecast for infrastructure and financial planning.

\subsubsection{Time Series Characteristics}\label{appendix:ts-character}
The selected common time series features~\cite{Hyndman2020Time1.0.2} are summarized as follows: 

\begin{enumerate}[-,topsep=4pt,parsep=0pt,leftmargin=24pt]
\item \textit{Trend:}  Trend component.
\item \textit{Spike:} `Spikiness';  variance of the remainder component.
\item \textit{Linearity:}  Based on coefficients of orthogonal quadratic regression.
\item \textit{Curvature:}  Based on coefficients of orthogonal quadratic regression.
\item \textit{Stability:}  Variance of the means of non-overlapping windows.
\item \textit{Lumpiness:}  Variance of the variance of non-overlapping windows.
\item \textit{Seasonal Strength:}  Seasonality component.
\item \textit{Trough:} Size and location of the troughs in the seasonal component.
\item \textit{Entropy:} `Shannon entropy'; measures the `forecast-ability' of a time series.
\item \textit{ACF1 and ACF10:}  First autocorrelation coefficient and sum of the first ten squared autocorrelation coefficients, respectively. Operating on the remainder part of the time series after subtracting trend and seasonality.
\item \textit{Peak:} Size and location of the peaks in the seasonal component.
\end{enumerate}

\subsubsection{Error Metrics}\label{appendix:metrics-def}
Given \( y_t \) as the actual value and \( \hat{y}_t \) as the predicted value at time \( t \), the error metrics for the period \(t \in [1, n]\) are defined as follows:
\begin{align}
    \text{RMSE}  &= \sqrt{\frac{1}{n} \sum_{t=1}^{n} (y_t - \hat{y}_t)^2}, \\
    \text{MAE}   &= \frac{1}{n} \sum_{t=1}^{n} |y_t - \hat{y}_t|, \\
    \text{MAPE}  &= \frac{100}{n} \sum_{t=1}^{n} \left| \frac{y_t - \hat{y}_t}{y_t} \right|, \\
    \text{MASE}  &= \frac{\sum_{t=1}^{n} |y_t - \hat{y}_t|}{\sum_{t=1}^{n} |y_t - y_{t-48}|}, \\
    \text{MSSE}  &= \frac{\sum_{t=1}^{n} (y_t - \hat{y}_t)^2}{\sum_{t=1}^{n} (y_t - y_{t-48})^2}
\end{align}
The denominators in MASE and MSSE correspond to the seasonal naive forecast, \( \hat{y}_{t}^{\mathrm{NAIVE}}=y_{t-48} \), utilizing the actual values from 48 hours prior, as the typically used data from 24 hours before is not accessible at the time of forecasting. 

\clearpage
\section{Supplementary Material on Temperature and Seasonal Patterns} 
\label{appendix-temp}
Supplementary material for manuscript 
"Extending Load Forecasting from Zonal Aggregates to Individual Nodes for Transmission System Operators``
submitted to ``Sustainable Energy, Grids and Networks'' 
by Oskar Triebe et al.  (2025).

\subsection{Dataset Seasonality Patterns}
\begin{figure}[htb]
\centering
\includegraphics[width=0.9\textwidth]{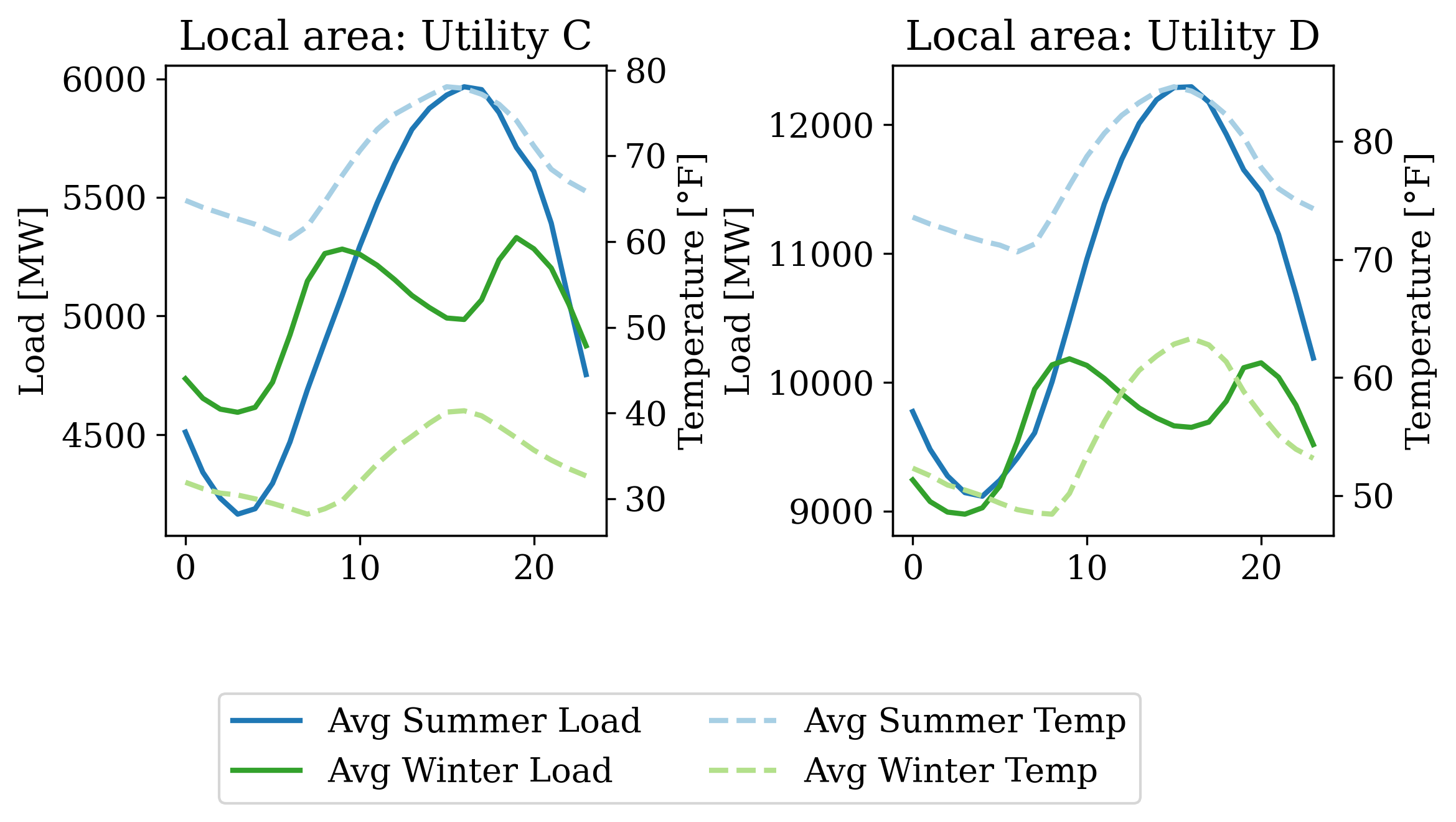}
\caption{Average daily load and temperature patterns in summer and winter for representative northern (left) and southern (right) utilities. 
}
\label{fig:seasonal_load}
\end{figure}

There is a strong relationship between temperature and load in our data. For example, the daily net load profile follows distinct seasonal patterns for summer and winter, illustrated in Figure \ref{fig:seasonal_load}. Summer load exhibits a pronounced mid-day peak. Winter load patterns are more nuanced, with distinct morning and evening peaks and a mid-day drop.
Compared to northern utility C, the southern utility D exhibits a larger summer winter difference, attributable to lower heating needs due to mild temperatures in winter, in contrast to high cooling demand in summer.

\newpage
\subsection{Model Seasonality Component}
\begin{figure}[htb]
\centering
\includegraphics[width=0.9\textwidth]{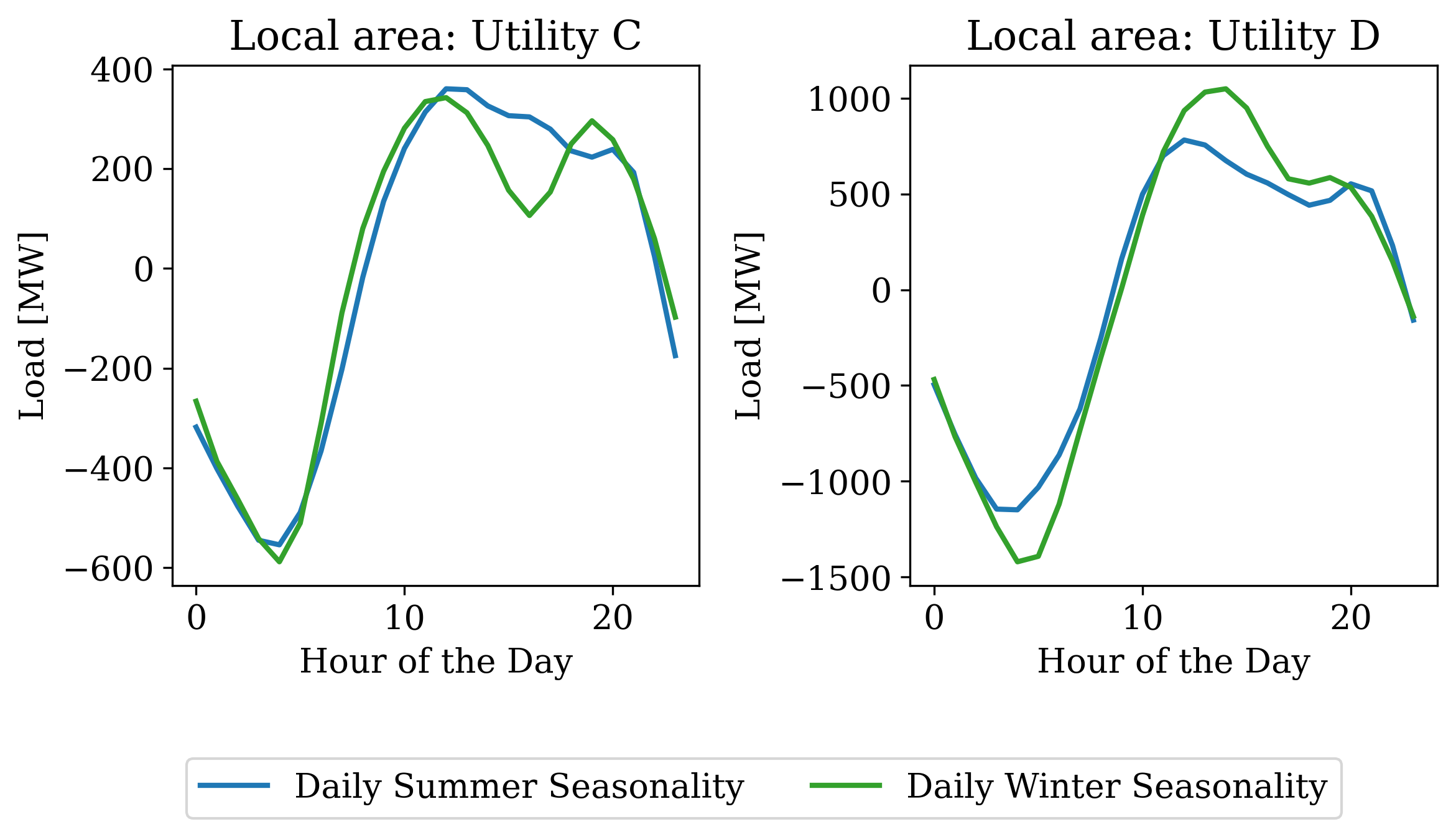} 
\caption{Daily seasonality model component for summer and winter, illustrated for representative northern (left) and southern (right) utilities.}
\label{fig:daily_season}
\end{figure}

The model's fitted daily seasonalities, presented in Figure \ref{fig:daily_season}, share some of the distinct mean load patterns observed during summer and winter seen in Figure \ref{fig:seasonal_load}. The winter seasonalities exhibit a decline in load during the afternoon hours, a feature that is  absent in the summer. 
Note that the fitted seasonalities are not expected to directly align with the mean data patterns in Figure \ref{fig:seasonal_load}, as related patterns are captured by other modeling components, such as autoregression or temperature regression.

\newpage
\subsection{Dataset Temperature-Load Relationship} 
\begin{figure}[htb]
\centering
\includegraphics[width=1.0\textwidth]{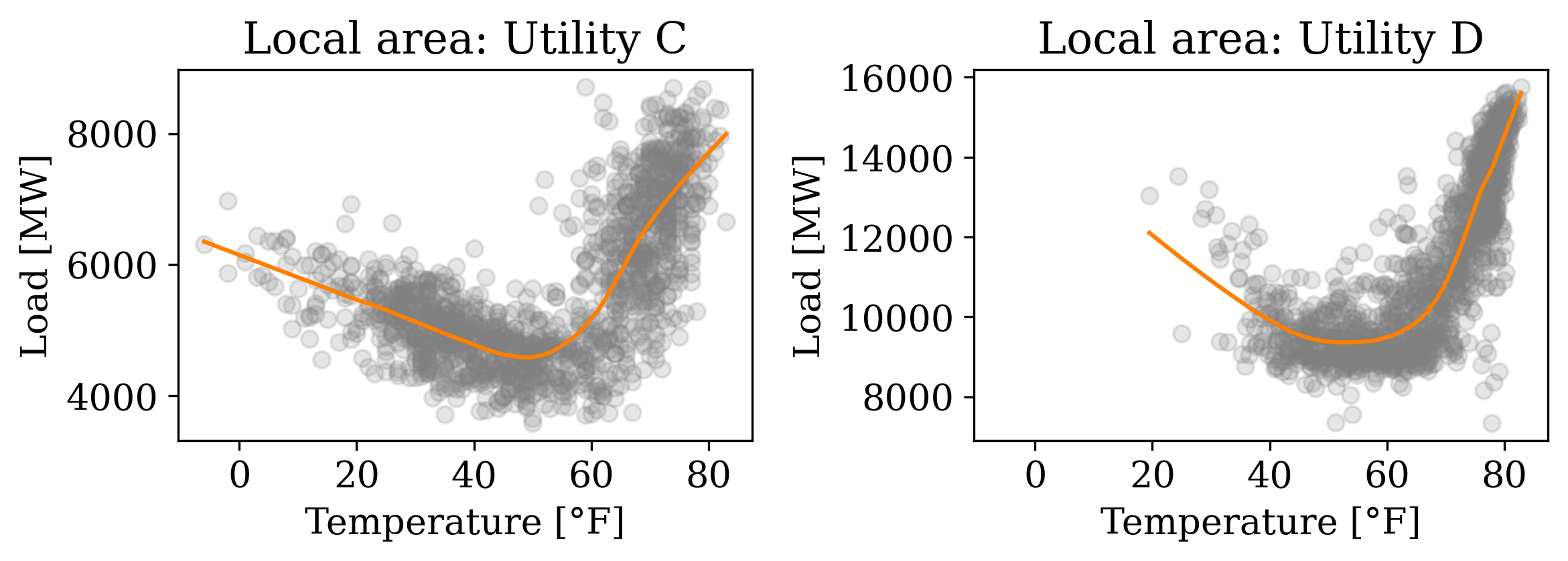}
\caption{Temperature-dependent variability in hourly load for representative northern (left) and southern (right) utilities.  }
\label{fig:load_temp_dependency}
\end{figure}

A V-shaped relationship emerges between temperature and load, illustrated in Figure \ref{fig:load_temp_dependency}. As temperatures drop below 57°F, load increases due to heightened heating needs. Conversely, above 57°F load also increases, signifying greater cooling demands.

\newpage
\subsection{Model Temperature Component}
\begin{figure}[htbp]
\centering
\includegraphics[width=1.0\textwidth]{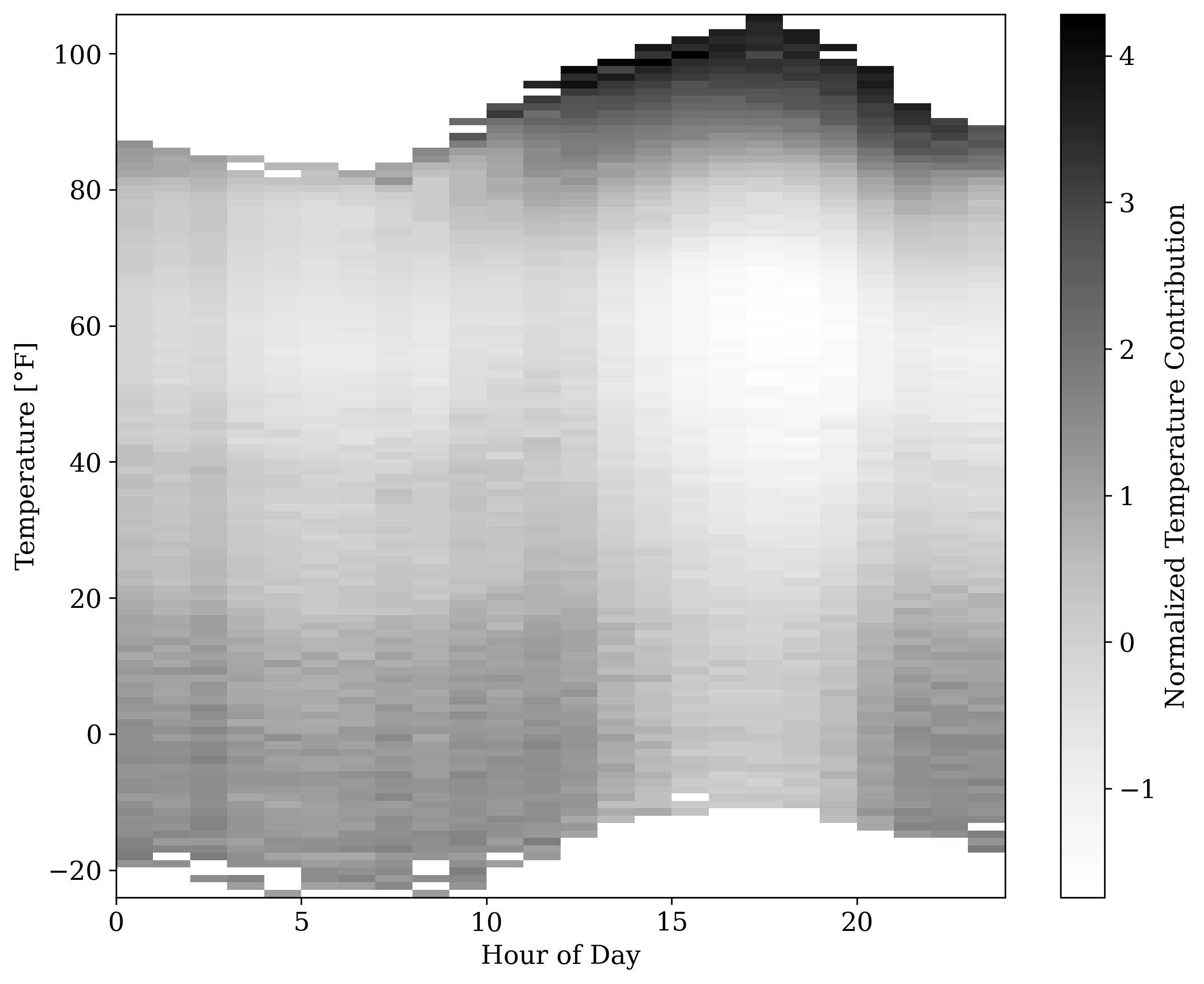}
\caption{The temperature component's normalized forecast contribution (relative to the other forecast components), averaged for binned temperature and hour of the day. Higher (darker) values signify an increase in the base forecast, lower (brighter) values indicate a decrease. Blank fields represent areas for which no data was recorded.}
\label{fig:temp_contribution}
\end{figure}

The temperature regression component of the model captures the distinct impacts of heating and cooling temperature areas, visualized in Figure \ref{fig:temp_contribution}. The inflection temperature is more nuanced -- varying for different hours of the day -- than the observed V-shape of the average load-temperature relationship in Figure \ref{fig:load_temp_dependency}.  
The temperature component impacts the forecast most during high daytime temperatures due to cooling demands. On the other extreme of the temperature spectrum, low night-time, morning and evening temperatures are also adding to the load due to heating demands. 

\clearpage
\section{Supplementary Material on Results}\label{appendix-results}
Supplementary material for manuscript 
"Extending Load Forecasting from Zonal Aggregates to Individual Nodes for Transmission System Operators``
submitted to ``Sustainable Energy, Grids and Networks'' 
by Oskar Triebe et al. (2025).

\subsection{Utility-level Accuracy of Scaled Aggregated Bus-Level Forecasts Compared to Actual Utility Load}\label{sec:appendix-case-utility-level}
The actual utility load differs from the aggregate of bus loads due to data source differences.
To demonstrate the worst-case performance of the studied approaches, we include an analysis using the actual utility load series as ground truth. 
Hereby, the `Global-Utility' model is identical to the global hits-GAM model evaluated in the utility-level forecasting study, while Local-Utility is a hits-GAM model locally fitted to Utility A only. Both utility-level methods are fitted on the actual utility load series. 

The bottom-up approaches `Global-Bus' and `Grouped Global-Bus' are disadvantaged in this comparison due to the data difference. They are fitted on bus-level data and trained to forecast individual bus loads, which are then summed up and scaled by their historical magnitudal relationship to the utility load series.
However, this aggregated bus load series, even if perfectly forecasted, does not match the utility series, as they have differing data sources.

\begin{table}[!htbp]
\centering
\caption{Utility-Level Forecast Accuracy (Utility Load)}
\label{tab:appendix-case-utility}
\small
\begin{tabular}{lrrrr}
\hline
Model & RMSE & MAE & MASE & MSSE \\
\hline
Global-Utility & 26.69 & 15.60 & 1.02 & 1.00 \\
Local-Utility & \textbf{19.39} & 13.83 & 0.91 & 0.80 \\
Global-Bus & 20.56 & 13.74 & 0.77 & 0.68 \\
Grouped Global-Bus & 19.91 & \textbf{13.18} &  \textbf{0.74} & \textbf{0.64}\\
\hline
\end{tabular}
\end{table}

Table~\ref{tab:appendix-case-utility} shows the forecast accuracy at utility-level for Utility A, using the actual utility load series as ground truth. 
Despite its data disadvantage, the `Global-Bus' model improves over the `Local-Utility' model on scaled metrics (MASE and MSSE), but performs similar by absolute metrics (MAE and RMSE), as seen in Table~\ref{tab:appendix-case-utility}. 
The `Grouped Global-Bus' model further improves upon this and performs the best across most metrics. 

The main study's bus level extension compares the utility-level forecast accuracy on the aggregate bus load instead of the actual utility load. There, all models are trained on the same data source and no scaling is needed. Thus, the metrics reported in the main study are more useful for comparing different methods and directly relatable to bus-level errors.

\subsection{Relationship to Time Series Characteristics}
\begin{figure}[!htbp]
\centering
\includegraphics[width=0.7\textwidth]{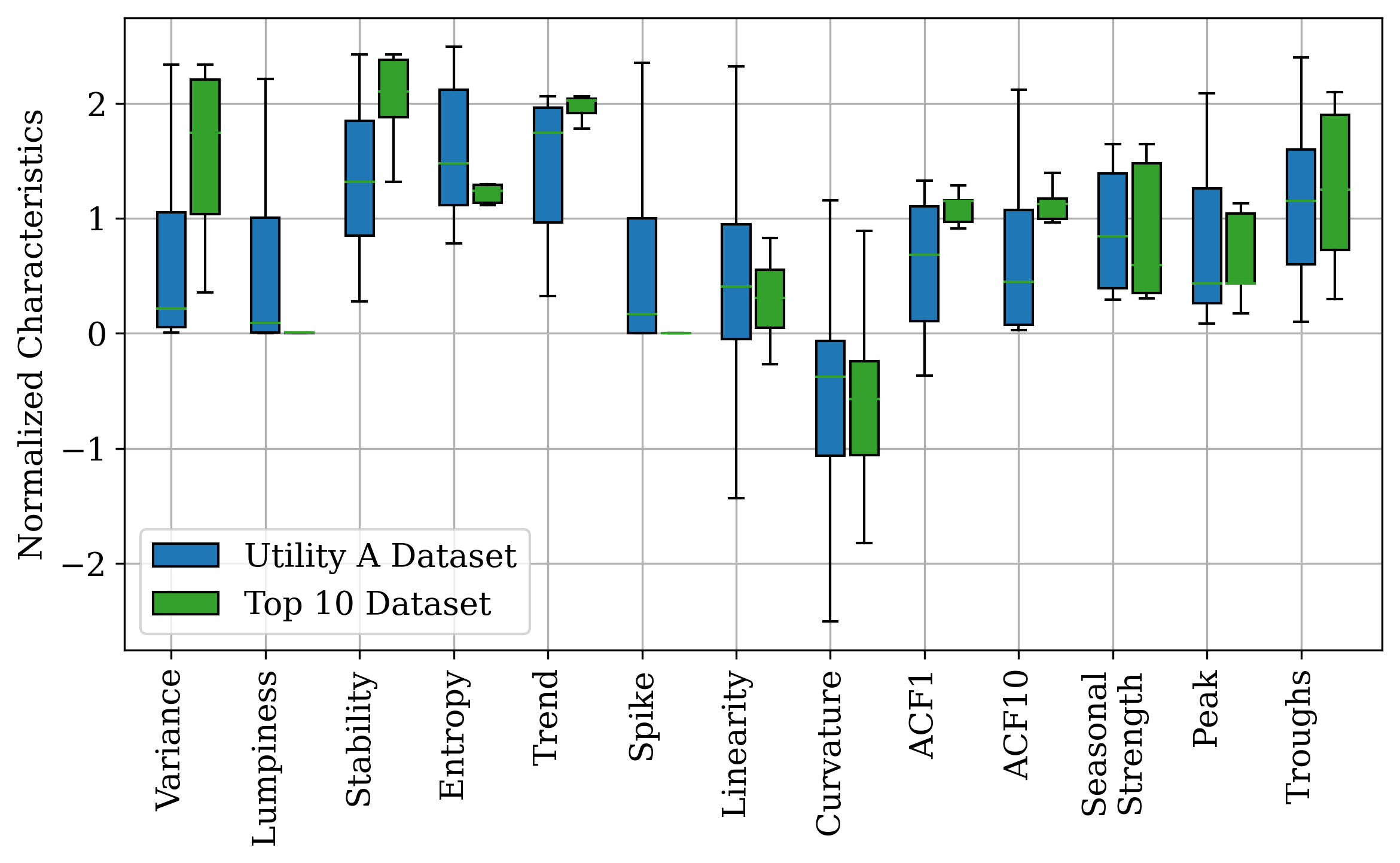}
\caption{Time series characteristics for all buses of Utility A (blue) and for the subset of the ten least accurate buses (green). Each characteristic is scaled by the interquartile range of all buses (blue box). 
}
\label{fig:case-top5-charcter}
\end{figure}

We further examine the time series characteristics of the 10 least accurate buses by forecasting error (MAE) in Figure~\ref{fig:case-top5-charcter}. Notably, the interquartile range of high-error buses (green box) clearly diverges from other buses (blue box) for characteristics such as variance, stability, ACF1, ACF10, lumpiness and spike. 
This suggests that the bus-level time series characteristics are related to forecast errors.

\subsection{Detailed Error Metrics}
Table \ref{tab:detailed_errors_utility} provides a detailed breakdown of error metrics for each utility, comparing the performance of different models. The table displays the minimum, first quartile (1Q), median, third quartile (3Q), maximum, and mean error values over all utilities. Table \ref{tab:detailed_errors_bus} shows the analogous breakdown for the bus level extension.

\begin{table*}[htbp]
\caption{Detailed error metrics on utility level \label{tab:detailed_errors_utility}}
\centering
\small
\begin{tabular}{llrrrrrr}
\hline
 & Model & Min & 1Q & Median & 3Q & Max & Mean \\ \hline
\multirow{5}{*}{RMSE {[}MW{]}} & ARIMA & 6.93 & 29.86 & 119.25 & 271.96 & 826.97 & 205.90 \\
 & kNN & 6.64 & 24.17 & 114.99 & 273.02 & 584.08 & 156.72 \\
 & sNa\"ive & 10.44 & 39.11 & 170.18 & 382.95 & 922.10 & 251.09 \\
 & XGBoost & 5.38 & 25.09 & 98.72 & 210.79 & 921.22 & 144.92 \\
 & hits-GAM & 4.10 & 21.69 & 73.82 & 152.06 & 519.27 & 105.41 \\ \hline
\multirow{5}{*}{MAE {[}MW{]}} & ARIMA & 5.49 & 23.26 & 92.68 & 205.26 & 608.04 & 155.53 \\
 & kNN & 5.19 & 18.16 & 84.32 & 196.21 & 420.55 & 114.30 \\
 & sNa\"ive & 8.20 & 30.32 & 134.98 & 293.51 & 660.01 & 189.84 \\
 & XGBoost & 4.10 & 19.15 & 72.82 & 159.04 & 724.42 & 108.61 \\
 & hits-GAM & 3.19 & 15.80 & 52.59 & 114.80 & 359.35 & 78.22 \\ \hline
\multirow{5}{*}{MAPE {[}\%{]}} & ARIMA & 2.94 & 6.08 & 7.57 & 8.99 & 10.61 & 7.35 \\
 & kNN & 2.39 & 5.12 & 5.73 & 6.81 & 8.33 & 5.76 \\
 & sNa\"ive & 3.64 & 8.44 & 9.60 & 11.05 & 12.81 & 9.46 \\
 & XGBoost & 2.35 & 4.51 & 4.94 & 5.87 & 12.79 & 5.53 \\
 & hits-GAM & 2.07 & 3.27 & 4.08 & 4.83 & 7.35 & 4.21 \\ \hline
\multirow{5}{*}{MASE {[}\%{]}} & ARIMA & 0.67 & 0.72 & 0.75 & 0.80 & 0.94 & 0.77 \\
 & kNN & 0.55 & 0.57 & 0.62 & 0.64 & 0.73 & 0.61 \\
 & sNa\"ive & 1.00 & 1.00 & 1.00 & 1.00 & 1.00 & 1.00 \\
 & XGBoost & 0.40 & 0.49 & 0.54 & 0.63 & 1.10 & 0.59 \\
 & hits-GAM & 0.38 & 0.50 & 0.56 & 0.64 & 1.02 & 0.58 \\ \hline
\multirow{5}{*}{MSSE {[}\%{]}} & ARIMA & 0.44 & 0.54 & 0.56 & 0.61 & 0.92 & 0.59 \\
 & kNN & 0.31 & 0.35 & 0.39 & 0.43 & 0.56 & 0.40 \\
 & sNa\"ive & 1.00 & 1.00 & 1.00 & 1.00 & 1.00 & 1.00 \\
 & XGBoost & 0.18 & 0.26 & 0.30 & 0.41 & 1.04 & 0.37 \\
 & hits-GAM & 0.16 & 0.25 & 0.31 & 0.40 & 1.00 & 0.36 \\
 \hline
\end{tabular}
\end{table*}

\begin{table*}[htbp]
\caption{Detailed error metrics on bus level }
\label{tab:detailed_errors_bus}
\centering
\small
\begin{tabular}{lrrrrrr}
\hline
RMSE {[}MW{]} & Min & 1Q & Median & 3Q & Max & Mean \\  \hline
Local-Utility & 0.28 & 1.53 & 3.39 & 5.07 & 76.70 & 7.88 \\
Global-Bus & 0.03 & 0.08 & 0.18 & 0.43 & 5.91 & 0.58 \\
Grouped Global-Bus & 0.03 & 0.08 & 0.18 & 0.41 & 6.05 & 0.59 \\ \hline
MAE {[}MW{]} & Min & 1Q & Median & 3Q & Max & Mean \\  \hline
Local-Utility & 0.27 & 1.52 & 3.36 & 5.02 & 76.11 & 7.79 \\
Global-Bus & 0.02 & 0.06 & 0.13 & 0.25 & 4.33 & 0.38 \\
Grouped Global-Bus & 0.02 & 0.06 & 0.12 & 0.25 & 4.40 & 0.39 \\ \hline
MAPE {[}\%{]} & Min & 1Q & Median & 3Q & Max & Mean \\  \hline
Local-Utility & 155.12 & 202.89 & 207.74 & 213.86 & 2931.52 & 253.62 \\
Global-Bus & 3.84 & 5.96 & 7.96 & 12.94 & 270.97 & 20.40 \\
Grouped Global-Bus & 3.93 & 5.76 & 8.23 & 14.10 & 268.87 & 21.02 \\ \hline
MASE {[}\%{]} & Min & 1Q & Median & 3Q & Max & Mean \\  \hline
Local-Utility & 1.49 & 2.00 & 2.04 & 2.08 & 2.94 & 2.06 \\
Global-Bus & 0.54 & 0.72 & 0.79 & 0.93 & 3.04 & 0.86 \\
Grouped Global-Bus & 0.53 & 0.71 & 0.81 & 0.97 & 2.93 & 0.87 \\ \hline
MSSE {[}\%{]}& Min & 1Q & Median & 3Q & Max & Mean \\  \hline
Local-Utility & 2.05 & 3.86 & 3.99 & 4.15 & 7.58 & 4.04 \\
Global-Bus & 0.32 & 0.53 & 0.63 & 0.75 & 1.86 & 0.67 \\
Grouped Global-Bus & 0.31 & 0.50 & 0.64 & 0.76 & 5.02 & 0.74 \\
 \hline
\end{tabular}
\end{table*}

\end{document}